\documentclass[10pt,twocolumn,letterpaper]{article}

\usepackage[pagenumbers]{cvpr} %

\usepackage{float}
\usepackage{array}  %

\definecolor{cvprblue}{rgb}{0.21,0.49,0.74}
\usepackage[breaklinks,colorlinks,allcolors=cvprblue]{hyperref}
\usepackage{fontawesome5} %

\def\confName{CVPR}
\def\confYear{2027}

\title{EarthRecord: From Pixels to Land Change with Multimodal Agents\\over Four Decades of Satellite Imagery}

\author{Haoran Zhang\textsuperscript{1,2} \quad Huizhen Sun\textsuperscript{1} \quad Hanbin Wen\textsuperscript{3} \quad Jie Leng\textsuperscript{4} \quad Jiaxin Zhu\textsuperscript{4} \quad Zhaohua Sun\textsuperscript{5,*}\\[0.5em]
{\small \textsuperscript{1}AIEyes Technologies (Hong Kong) Co., Limited, Hong Kong SAR, China}\\
{\small \textsuperscript{2}College of Future Technology, The Hong Kong University of Science and Technology (Guangzhou), Guangzhou 511400, China}\\
{\small \textsuperscript{3}School of Physics and Optoelectronic Engineering, Guangdong University of Technology, Guangzhou 510006, China}\\
{\small \textsuperscript{4}Anhua Ocean Intelligent Equipment Co., Ltd., Shenzhen 518000, China}\\
{\small \textsuperscript{5}Department of Mechanical Engineering, The University of Hong Kong, Hong Kong SAR, China}\\
{\small \textsuperscript{*}Corresponding author: Zhaohua Sun ({\tt joshsun@hku.hk})}\\[0.7em]
{\normalsize\renewcommand{\arraystretch}{1.15}\begin{tabular}{@{}l@{}}
\makebox[1.5em][l]{\raisebox{-0.2em}{\includegraphics[height=1.2em]{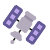}}}\textbf{Project Page:} \url{https://ranranrannervous.github.io/EarthRecord/}\\
\makebox[1.5em][l]{\faGithub}\textbf{Code:} \url{https://github.com/ranranrannervous/EarthRecord}\\
\makebox[1.5em][l]{\raisebox{-0.12em}{\includegraphics[height=1em]{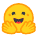}}}\textbf{Dataset:} \url{https://huggingface.co/datasets/haoranranranran/EarthRecord}
\end{tabular}}
}

\begin{document}
\twocolumn[{%
  \renewcommand\twocolumn[1][]{#1}%
  \maketitle
  \begin{center}
    \captionsetup{type=figure}
    \includegraphics[width=\linewidth]{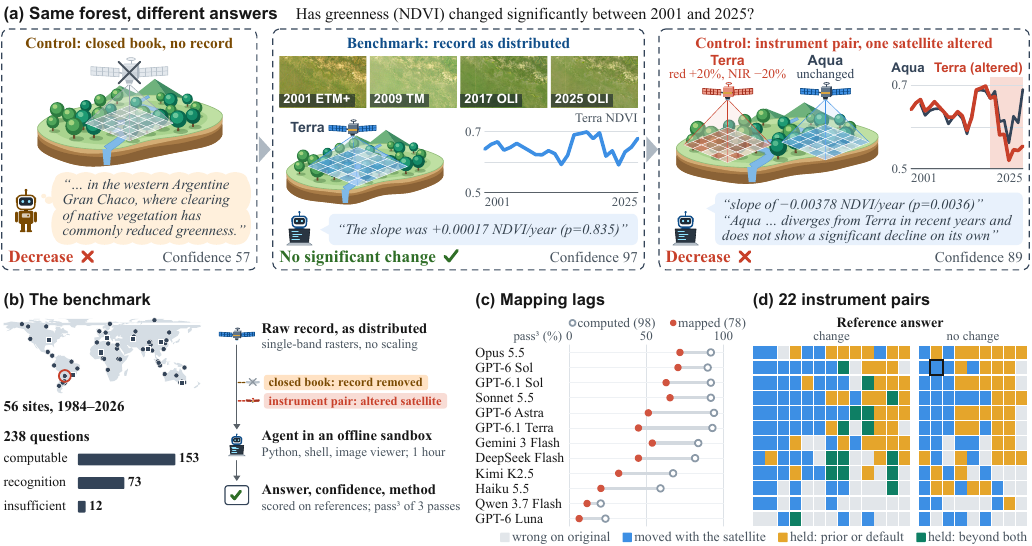}
    \caption{\textbf{Overview of EarthRecord.}
    (a) One question answered by GPT-6 Sol under the three conditions (quotes from one of three runs). Without the record, it answers from regional priors; with the record, it computes the right answer from the pixels; with one satellite altered, it follows the altered pixels although the forest has not changed.
    (b) Benchmark and protocol.
    (c, d) Twelve agents: answers needing a map lag those computed from the record, and half of right answers change when one satellite's data are altered.
    Red circle in (b) and black box in (d) mark the site and case of (a).
    Landsat data courtesy of the U.S. Geological Survey.}
    \label{fig:ladder}
  \end{center}
}]
\begin{abstract}
While multimodal agents now carry out complex data analysis with tools, their ability to turn decades of satellite imagery into reliable land-change analysis, crucial for Earth observation science, remains unmeasured.
Existing benchmarks fall short: they pose questions on curated images or give agents preprocessed inputs, leaving out the raw multi-sensor archives where such analysis begins.
We introduce EarthRecord, a benchmark built on the raw record of 56 sites: four decades of Landsat, MODIS, VIIRS, and Sentinel-2 data, with cloud masking and cross-sensor harmonization left to the agent.
It poses 238 land-change questions across ten analysis tasks, scored against withheld references from global products, agreement among methods, and visual interpretation.
Two controls test what answers rest on: the closed book removes the record, and instrument pairs alter one satellite's data while nothing changes on the ground.
Evaluations of twelve multimodal agents reveal that leading agents complete most analyses end to end but falter when an answer requires mapping from imagery.
Further analysis shows that the record moves answers from prior knowledge to the pixels but not yet to the real change on the ground: when one satellite's data are altered, half of the right answers follow the alteration.
EarthRecord thus provides a principled platform for measuring and advancing agents from pixels to land change.
\end{abstract}

\begin{table*}[t]
  \centering
  \footnotesize
  \newcommand{\yes}{\textcolor{OliveGreen}{\checkmark}}
  \newcommand{\no}{\textcolor{BrickRed}{$\times$}}
  \newcommand{\hd}[1]{\begin{tabular}[b]{@{}c@{}}#1\end{tabular}}
  \setlength{\tabcolsep}{8pt}
  \begin{tabular}{llccccc}
    \toprule
    & & \multicolumn{3}{c}{Setting and data} & \multicolumn{2}{c}{Controls} \\
    \cmidrule(lr){3-5} \cmidrule(lr){6-7}
    Group & Benchmark & Agent & \hd{Multi-decadal,\\multi-sensor} & Raw record & Prior & Sensor / ground \\
    \midrule
    Vision-language & MMStar~\cite{chen2024mmstar} & \no & \no & \no & \yes & \no \\
    & NaturalBench~\cite{li2024naturalbench} & \no & \no & \no & \no & \no \\
    & OSI-Bench~\cite{wu2026from} & \no & \no & \no & \yes & \no \\
    \midrule
    Remote sensing vision-language & CHOICE~\cite{an2025choice} & \no & \no & \no & \yes & \no \\
    & DVL-Bench~\cite{xuan2025dynamicvl} & \no & \no & \no & \no & \no \\
    & GeoMMBench~\cite{xiao2026geommbench} & \no & \no & \no & \yes & \no \\
    \midrule
    Remote sensing foundation models & RSPDBench~\cite{faruk2027rspdbench} & \no & \no & \no & \no & \yes \\
    & EarthShift~\cite{doerksen2026earthshift} & \no & \no & \no & \no & \yes \\
    \midrule
    Earth observation agents & UnivEARTH~\cite{kao2026towards} & \yes & \yes & \no & \yes & \no \\
    & Earth-Bench~\cite{feng2026earth} & \yes & \no & \no & \yes & \no \\
    & ThinkGeo~\cite{shabbir2026thinkgeo} & \yes & \no & \no & \no & \no \\
    & TerraBench~\cite{nguyen2026terrabench} & \yes & \no & \no & \no & \no \\
    & DORA~\cite{wang2026can} & \yes & \no & \no & \no & \no \\
    & AgenticCity-Bench~\cite{yu2026agenticcity} & \yes & \yes & \no & \no & \no \\
    \midrule
    & \textbf{EarthRecord (ours)} & \yes & \yes & \yes & \yes & \yes \\
    \bottomrule
  \end{tabular}
  \caption{\textbf{Comparison of benchmarks.} Only EarthRecord tests agents on the raw, multi-decadal, multi-sensor record with both controls.}
  \label{tab:benchmarks}
\end{table*}

\section{Introduction}
\label{sec:intro}

Multimodal agents now open files, write code, and carry out multi-step data analysis with tools~\cite{chen2025scienceagentbench,jimenez2024swebench,jing2025dsbench,lei2025spider,yao2023react}.
Turning the Earth's fifty-year satellite record~\cite{crawford2023landsat,wulder2022fifty} into reliable land-change analysis---tracking forest loss, surface water, settlements, and fire~\cite{hansen2013high,pekel2016high,pesaresi2023ghs,potapov2022global,giglio2021mcd64a1}---is both a central task of Earth observation science and a demanding test of these agents.
In practice, such analysis begins not with a clean RGB image, but with decades of observations distributed across multiple satellite missions, where an analyst must screen clouds and gaps, map transitions from imagery, and harmonize measurements across sensors and across years~\cite{claverie2018harmonized,markham2012forty,roy2016characterization,schaepman2006reflectance}.
Whether multimodal agents can carry out these analyses end to end remains unmeasured.

Existing benchmarks fall short in both the data they provide and what their scores verify (\cref{tab:benchmarks}).
Vision-language~\cite{chen2024mmstar,li2024naturalbench,wu2026from} and remote-sensing multimodal benchmarks~\cite{an2025choice,xiao2026geommbench,xuan2025dynamicvl} pose questions on curated images, and recent Earth-observation agent benchmarks~\cite{feng2026earth,kao2026towards,lv2026earthagentpro,nguyen2026terrabench,shabbir2026thinkgeo,wang2026can,yu2026agenticcity} still hand agents preprocessed inputs or short-term tasks, or let them fetch data through dedicated tools, leaving out the raw, multi-decadal, multi-sensor archives where land-change analysis begins.
Moreover, scoring the final answer alone does not verify how the agent reached it.
Models often answer about a region from \emph{prior knowledge} without using the data~\cite{agrawal2018dont,chen2024mmstar,li2024naturalbench,wu2024clasheval,xie2024adaptive}.
Even when an agent computes from the recorded \emph{pixels}, a sensor measures radiance that depends on the atmosphere, viewing geometry, and instrument as well as on the \emph{ground}~\cite{markham2012forty,schaepman2006reflectance}.
In the satellite record, sensor shifts have repeatedly created apparent changes on Earth: tropospheric cooling from orbital decay~\cite{wentz1998effects}, vegetation browning from a degrading sensor~\cite{wang2012impact,zhang2017reanalysis}, and an Amazon green-up from viewing geometry~\cite{morton2014amazon}.
Each time the pixel values moved, the question was whether the ground had changed~\cite{claverie2018harmonized,doerksen2026earthshift,faruk2027rspdbench,roy2016characterization}.

We introduce EarthRecord, a benchmark that evaluates whether multimodal agents can turn four decades of raw satellite records into reliable land-change analysis (\cref{fig:ladder}).
In an offline sandbox with Python, shell, and an image viewer, an agent receives the record of a site as distributed~\cite{crawford2023landsat}---unscaled single-band rasters from Landsat, MODIS, VIIRS, and Sentinel-2 spanning 1984--2026 across 56 sites---leaving cloud masking, mapping, and cross-sensor harmonization to the agent.
EarthRecord poses 238 questions across ten analysis tasks (153 computable from the record, 73 asking what is on the ground, and 12 with insufficient evidence), scored by pass$^3$ across three passes against withheld references from global products, agreement among methods, and visual interpretation.
Two controls test what answers rest on: the \emph{closed book} removes the record to measure what prior knowledge alone can answer, and \emph{instrument pairs} alter one satellite's data while nothing changes on the ground, testing cross-sensor checks.

Evaluating twelve multimodal agents on EarthRecord reveals both their end-to-end capability and two fundamental bottlenecks (\cref{fig:ladder}c, d).
First, the four leading agents complete most analyses end to end---scoring 81.0\% to 85.4\% pass$^3$ on the 226 computable and recognition questions---but the six leading models drop from 92.0\% on questions computed from record variables to 61.1\% on those requiring a map from imagery, where runs often abstain early, mostly without viewing an image, or judge inconsistently across passes (\cref{sec:experiments}).
Second, the controls show that the record moves answers from prior knowledge to the pixels, but not yet to the ground (\cref{sec:analysis}).
Providing the record lifts all twelve models on computable questions, though regional priors persist on recognition questions.
When one satellite's data are altered in instrument pairs, however, 53.5\% of right answers turn wrong and follow the altered pixels, and few failed runs attribute the break to the instrument.

Our contributions are three.
\begin{itemize}
  \item \textbf{A benchmark on the raw satellite record.}
  EarthRecord pairs four decades of unscaled Landsat, MODIS, VIIRS, and Sentinel-2 archives at 56 sites with 238 land-change questions across ten analysis tasks.
  \item \textbf{Controls for prior knowledge and sensor shifts.}
  Closed-book removal and altered instrument pairs test whether answers rest on regional priors, unverified pixels, or cross-checked change on the ground.
  \item \textbf{An evaluation of twelve multimodal agents.}
  We show that leading agents complete most variable-computed analyses end to end but falter at mapping from imagery, and half of right answers follow altered sensors.
\end{itemize}

\begin{figure*}[t]
  \centering
  \includegraphics[width=\linewidth]{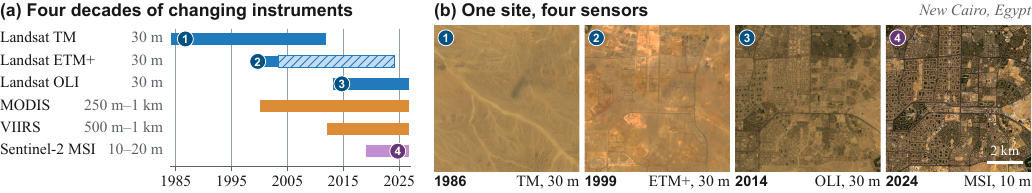}
  \caption{\textbf{The record spans changing instruments.}
  (a) Sensor coverage in the record, with hatching after the 2003 Landsat 7 scan-line corrector failure.
  (b) True-color crops of one site on the dates marked 1--4, with the same extent and reflectance stretch.
  Contains modified Copernicus Sentinel data 2024. Landsat data courtesy of the U.S. Geological Survey.}
  \label{fig:record}
\end{figure*}

\section{EarthRecord}
\label{sec:earthrecord}

Each EarthRecord question asks about the land surface at a given site over a given period.
In an offline sandbox (\cref{sec:protocol} and \cref{sec:supp-models}), the agent receives the site's \emph{raw record} as distributed by its providers---leaving data selection, cloud masking, scaling, mapping, and cross-sensor harmonization to the agent---and submits a structured answer scored against verifiable references under the standard evaluation and two controls.

\subsection{Sites and the raw record}
\label{sec:sites}

EarthRecord spans 56 sites: 48 core areas of 10--30\,km on a side and 8 administrative regions~\cite{runfola2020geoboundaries}.
The core areas cover eight themes, with six areas each, across six continents and all five K\"{o}ppen climate groups~\cite{beck2023koppen}, selected by screening a global grid for thematic signal and then reviewed manually.
To probe how prior knowledge interacts with the data, they include 8 \emph{famous} areas, 7 \emph{stable} areas selected for their lack of significant change, and 33 \emph{obscure} areas; the agent can read each area's coordinates.

The record combines four decades of observations up to its snapshot date of 30 September 2026 (\cref{fig:record}).
Landsat surface reflectance begins in 1984, spans TM, ETM+, and OLI~\cite{crawford2023landsat}, and provides a median of 1,101 scenes per core area; 109 of the 238 questions span two or more of these Landsat sensors (\cref{fig:record}a).
MODIS vegetation indices, land surface temperature, and snow cover begin in 2000 on Terra and later Aqua~\cite{didan2021mod13q1,didan2021myd13q1,didan2021mod13a1,didan2021myd13a1,wan2021mod11a2,wan2021myd11a2,hall2021mod10a1,hall2021myd10a1}, and VIIRS vegetation indices and temperature begin in 2012~\cite{vermote2023vnp13a1,hulley2023vnp21a2}.
Sentinel-2 surface reflectance enters in 2019 with at most one scene per month~\cite{drusch2012sentinel}, and processing baselines from 2022 add a radiometric offset~\cite{copernicus2026s2processing}.
ERA5-Land reanalysis~\cite{c3s2019era5land,munozsabater2021era5land}, global land cover~\cite{friedl2022mcd12q1,zanaga2021worldcover,zanaga2022worldcover}, terrain~\cite{esa2022copernicusdem}, and administrative boundaries complete the record, and the regions have no Landsat or Sentinel-2 data.

Apart from the regional boundary mask and the instrument pairs (\cref{sec:controls}), pixel values are unmodified: scale factors, offsets, and quality layers are not applied, and a README gives the conversions.
Duplicate and at least 90\% obscured scenes are removed, and archive issues present as distributed are kept: of the 57,157 Landsat scenes over the core areas, 4,485 are Tier 2~\cite{usgs2026tiers}, with geometric offsets of up to 15.7\,km, and quality layers missed cloud in 5 of 200 sampled scenes they rated clear.

\subsection{Questions and analysis tasks}
\label{sec:questions}

The benchmark poses 238 questions---207 at core areas across eight themes (forest, greening, agriculture, water, urban, fire, snow, and coast) and 31 at the regions---over periods of 1 to 38 years.
Every question specifies structured answer fields, including a designated conclusion field used for scoring, and every field accepts \emph{insufficient evidence}.
Each question belongs to one of ten analysis tasks set by its template (\cref{sec:supp-tasks}): \emph{Quantify} (49), \emph{Trend} (28), \emph{Date} (26), \emph{Compare} (24), \emph{Regional statistics} (21), \emph{Attribute} (20), \emph{Anomaly and climate} (19), \emph{Detect} (16), \emph{Season} (12), and \emph{Recovery} (11), counted over the 226 computable and recognition questions.

\begin{figure*}[t]
  \centering
  \includegraphics[width=\linewidth]{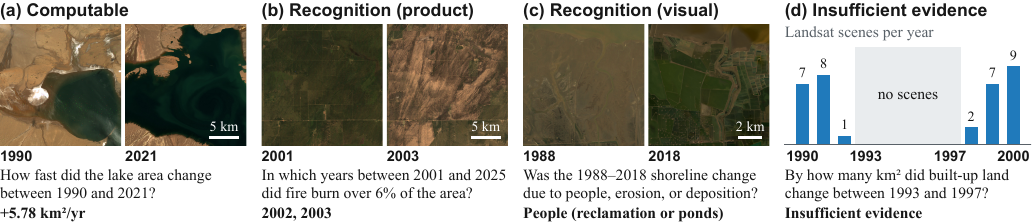}
  \caption{\textbf{One example question per class.}
  Questions are shortened, and bold text gives the reference answer.
  (a--c) Two Landsat scenes of one site, with the same extent and reflectance stretch within each pair.
  (d) Landsat scenes per year at one site, with the question period in gray.
  Landsat data courtesy of the U.S. Geological Survey.}
  \label{fig:cases}
\end{figure*}

Questions fall into three classes, assigned from the target variable.
\emph{Computable} questions (153, including all 31 regional questions) ask for quantities that a formula can derive from the record, such as greenness trends, lake-area trends, or snow-cover days (\cref{fig:cases}a).
\emph{Recognition} questions (73, of which 50 are in the forest, fire, and urban themes) ask what is on the ground or what happened, such as forest loss, burn scars, built-up expansion, or coastal reclamation (41 with product references and 32 with visual references; \cref{fig:cases}b, c).
\emph{Insufficient-evidence} questions (12) cover periods with too few clear observations to support a conclusion (\cref{fig:cases}d).

\subsection{Reference answers}
\label{sec:references}

The target quantity determines the source of each reference answer.
First, the 98 computable questions without a global product take references derived from variables in the record: several methods compute each quantity, numeric references take the method median, and categorical and date conclusions require at least 75\% of methods to agree.
Reference methods use Tier 1 Landsat scenes with bright-cloud pixels removed, and the 66 questions that rely on Landsat methods reach the same conclusions with and without this step.
Second, 13 questions take their reference from global products included in the record.
Third, 78 questions take their reference from four global products \emph{withheld} from the record: Hansen forest loss~\cite{hansen2013high}, JRC surface water~\cite{pekel2016high}, GHSL built-up surface~\cite{pesaresi2023ghs}, and MCD64A1 burned area~\cite{giglio2021mcd64a1}.
To answer these 41 recognition and 37 computable questions (such as the lake-area trend in \cref{fig:cases}a), the agent must map the change from the imagery itself.
Fourth, causes and land-cover types, and cases where products and methods diverge widely, are set by human review of AI-drafted visual interpretations of the imagery (37 questions: 32 recognition and 5 computable).
The 12 insufficient-evidence questions are set by the number of observations and the clear-sky fraction in the queried period.

A numeric answer is right when it falls within the \emph{reference range}, which 159 questions derive from product and method results and which spans at least $\pm$5\% of the reference value.
Decision thresholds are stated in the question text, and questions with results close to a threshold were replaced during construction.

\subsection{Protocol and scoring}
\label{sec:protocol}

Each run uses a fresh, offline, CPU-only container with 4 CPUs and 24\,GB of memory, Python, a shell, geospatial and classical machine-learning libraries, and an image viewer.
Because the record is delivered as single-band raster files and the viewer reads rendered images, an agent must select bands, apply a stretch, and render an image before it can inspect it.
All models share the same system prompt and tools, each run is limited to 3,600\,s of working time and 300 messages, and each answer is submitted with a confidence and a method summary with caveats.

Every question is run for three passes and scored on its conclusion field by pass$^3$~\cite{yao2024tau}, which is 1 when the conclusion is right in all three passes and 0 otherwise.
Intervals are 95\% bootstrap intervals over questions, and the baselines are the majority class and an always-no-change answer, which is the reference answer for 32 questions.
The first model's run served as a pilot check of the bank: after its results, references were audited and corrected across the entire bank by question-type rules, 11 questions with unstable references were removed, and pass$^3$ was set as the main score.
Because question texts did not change, that model was rescored rather than rerun, and the other eleven models ran on the frozen bank.

\subsection{Controls}
\label{sec:controls}

Two controls test what the conclusions rest on.
Closed-book runs give the model the same question and the site's bounding box without the record, and measure what prior knowledge alone can answer; the difference from the score with the record is the gain from the record (\cref{tab:analysis}).

Instrument pairs alter one satellite's data in the record while nothing changes on the ground, and score the answer against the original question and reference answer.
All 22 pairs use computable questions: 21 scale one satellite's values from 2018 onward, by up to 20\% in 20 pairs and by 5\% in one, and the remaining pair replaces Landsat 8 scenes from 2013--2017 with Landsat 7 scenes.
Every retained pair is detectable by comparison with concurrent satellites, and the alteration changes the conclusion of a reference method that relies on the affected satellite.
Eight placebo pairs remove 5\% of clear scenes at random and measure conclusion variability unrelated to the instrument.
Model--pair cases right on the original in all three passes are sorted by the altered version: moved cases turn wrong, held cases stay right but are already right in the closed book or have no change as the reference, and beyond cases stay right otherwise (\cref{sec:pixels-world}).

\providecommand{\mlogo}[1]{\raisebox{-0.2em}{\includegraphics[height=1.05em]{figures/logos/#1.pdf}}\hspace{0.45em}}
\providecommand{\mlogoskip}{\hspace{1.5em}}
\begin{table*}[t]
  \centering
  \footnotesize
  \newcommand{\ci}[2]{\,{\scriptsize\textcolor{gray}{[#1, #2]}}}
  \newcommand{\hd}[1]{\begin{tabular}[b]{@{}c@{}}#1\end{tabular}}
  \setlength{\tabcolsep}{4pt}
  \renewcommand{\arraystretch}{1.05}
  \begin{tabular}{lccccccccccc}
    \toprule
    & & \multicolumn{10}{c}{Analysis task} \\
    \cmidrule(lr){3-12}
    Model & Overall & Quantify & Date & \hd{Anomaly and\\climate} & Detect & Trend & Recovery & Season & Compare & Attribute & \hd{Regional\\statistics} \\
    & \scriptsize(226) & \scriptsize(49) & \scriptsize(26) & \scriptsize(19) & \scriptsize(16) & \scriptsize(28) & \scriptsize(11) & \scriptsize(12) & \scriptsize(24) & \scriptsize(20) & \scriptsize(21) \\
    \midrule
    \mlogo{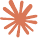}Opus 5.5 & \textbf{85.4}\ci{80.5}{89.8} & 69.4 & 53.8 & \textbf{84.2} & \textbf{100.0} & \textbf{96.4} & 81.8 & \textbf{100.0} & \textbf{100.0} & \textbf{100.0} & \textbf{100.0} \\
    \mlogo{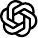}GPT-6 Sol & 84.1\ci{79.2}{88.9} & \textbf{75.5} & \textbf{65.4} & 78.9 & 87.5 & 89.3 & \textbf{100.0} & 91.7 & 79.2 & \textbf{100.0} & \textbf{100.0} \\
    \mlogo{openai}GPT-6.1 Sol & 81.4\ci{76.1}{86.3} & 63.3 & \textbf{65.4} & 73.7 & 81.2 & 89.3 & \textbf{100.0} & 83.3 & 91.7 & \textbf{100.0} & \textbf{100.0} \\
    \mlogo{claude}Sonnet 5.5 & 81.0\ci{75.7}{86.3} & 67.3 & 61.5 & 78.9 & 87.5 & 89.3 & 81.8 & 83.3 & 87.5 & 95.0 & \textbf{100.0} \\
    \mlogo{openai}GPT-6 Astra & 77.0\ci{71.2}{82.3} & 55.1 & 53.8 & 68.4 & 81.2 & 78.6 & \textbf{100.0} & 91.7 & 95.8 & \textbf{100.0} & 95.2 \\
    \mlogo{openai}GPT-6.1 Terra & 73.5\ci{67.7}{79.2} & 40.8 & 57.7 & 73.7 & 75.0 & 85.7 & 90.9 & 91.7 & 83.3 & 95.0 & \textbf{100.0} \\
    \mlogo{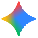}Gemini 3 Flash & 72.6\ci{66.8}{77.9} & 51.0 & 53.8 & 63.2 & 62.5 & 75.0 & 81.8 & \textbf{100.0} & 87.5 & \textbf{100.0} & 95.2 \\
    \mlogo{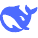}DeepSeek Flash & 68.1\ci{61.9}{74.3} & 49.0 & 46.2 & 68.4 & 68.8 & 75.0 & 63.6 & 75.0 & 87.5 & 80.0 & 95.2 \\
    \mlogo{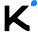}Kimi K2.5 & 54.0\ci{47.8}{60.6} & 38.8 & 46.2 & 36.8 & 56.2 & 75.0 & 72.7 & 58.3 & 58.3 & 60.0 & 61.9 \\
    \mlogo{claude}Haiku 5.5 & 41.2\ci{35.0}{47.8} & 24.5 & 15.4 & 42.1 & 25.0 & 60.7 & 36.4 & 83.3 & 50.0 & 50.0 & 57.1 \\
    \mlogo{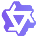}Qwen 3.7 Flash & 19.5\ci{14.6}{24.8} & 6.1 & 7.7 & 0.0 & 31.2 & 39.3 & 54.5 & 8.3 & 37.5 & 35.0 & 0.0 \\
    \mlogo{openai}GPT-6 Luna & 18.6\ci{13.7}{23.9} & 14.3 & 19.2 & 15.8 & 12.5 & 21.4 & 18.2 & 41.7 & 12.5 & 45.0 & 0.0 \\
    \midrule
    \mlogoskip\textit{Majority class} & 23.9 & 16.3 & 7.7 & 5.3 & 18.8 & 42.9 & 90.9 & 25.0 & 29.2 & 40.0 & 0.0 \\
    \mlogoskip\textit{Always no change} & 13.7 & 2.0 & 0.0 & 0.0 & 62.5 & 35.7 & 9.1 & 8.3 & 33.3 & 0.0 & 0.0 \\
    \bottomrule
  \end{tabular}
  \caption{\textbf{Results by analysis task.}
  Scores are pass$^3$ (\%) on the 226 computable and recognition questions, with question counts in parentheses and 95\% intervals for the overall score.
  Tasks run from the lowest to the highest median score, and rows are sorted by the overall score.
  \cref{sec:supp-tasks} shows an example of each task, and \cref{sec:supp-themes} splits the scores by class and theme.}
  \label{tab:main}
\end{table*}

\begin{table*}[t]
  \centering
  \footnotesize
  \newcommand{\hd}[1]{\begin{tabular}[b]{@{}c@{}}#1\end{tabular}}
  \setlength{\tabcolsep}{4.2pt}
  \renewcommand{\arraystretch}{1.05}
  \begin{tabular}{lcccccccccc}
    \toprule
    & \multicolumn{2}{c}{Accuracy} & Consistency & Confidence & \multicolumn{2}{c}{Abstention} & \multicolumn{3}{c}{Work per run} \\
    \cmidrule(lr){2-3} \cmidrule(lr){4-4} \cmidrule(lr){5-5} \cmidrule(lr){6-7} \cmidrule(lr){8-10}
    Model & One run & pass$^3$ & \hd{Right in one or\\two passes (of 226)} & AUROC & \hd{Recognition\\runs (\%)} & \hd{Insufficient-evidence\\runs (of 36)} & \hd{Tool\\calls} & Minutes & \hd{Image\\views} \\
    \midrule
    \mlogo{claude}Opus 5.5 & \textbf{88.8} & \textbf{85.4} & 13 & 0.79 & 4.2 & 35 & 9.1 & 4.7 & 0.59 \\
    \mlogo{openai}GPT-6 Sol & 88.6 & 84.1 & 20 & 0.76 & 4.6 & 35 & 9.0 & 2.4 & 0.31 \\
    \mlogo{openai}GPT-6.1 Sol & 83.9 & 81.4 & \textbf{11} & 0.79 & 29.7 & 36 & 8.6 & 2.8 & 0.52 \\
    \mlogo{claude}Sonnet 5.5 & 86.9 & 81.0 & 28 & 0.84 & 5.0 & 30 & 7.9 & 4.0 & 0.19 \\
    \mlogo{openai}GPT-6 Astra & 81.0 & 77.0 & 18 & 0.78 & 38.8 & 36 & 6.7 & 2.1 & 0.26 \\
    \mlogo{openai}GPT-6.1 Terra & 78.2 & 73.5 & 20 & \textbf{0.85} & 42.9 & 36 & 4.8 & 1.0 & 0.09 \\
    \mlogo{gemini}Gemini 3 Flash & 80.8 & 72.6 & 36 & 0.67 & 0.0 & 5 & 27.9 & 6.9 & 0.57 \\
    \mlogo{deepseek}DeepSeek Flash & 80.7 & 68.1 & 49 & 0.75 & 1.4 & 33 & 57.5 & 20.2 & 6.30 \\
    \mlogo{kimi}Kimi K2.5 & 68.7 & 54.0 & 62 & 0.70 & 4.1 & 20 & 32.4 & 10.9 & 0.58 \\
    \mlogo{claude}Haiku 5.5 & 62.1 & 41.2 & 83 & 0.70 & 32.0 & 32 & 6.3 & 3.1 & 0.03 \\
    \mlogo{qwen}Qwen 3.7 Flash & 31.1 & 19.5 & 53 & 0.63 & 39.7 & 33 & 47.5 & 16.9 & 0.08 \\
    \mlogo{openai}GPT-6 Luna & 40.4 & 18.6 & 100 & 0.67 & 40.2 & 33 & 6.0 & 0.9 & 0.01 \\
    \bottomrule
  \end{tabular}
  \caption{\textbf{Agent behavior and effort.}
  All columns cover the 226 computable and recognition questions, except the last abstention column, which counts runs on the 12 insufficient-evidence questions.
  Abstention counts runs that answer insufficient evidence, and AUROC rates how well reported confidence separates right from wrong conclusions.
  Bold marks the best model in each of the first four columns.}
  \label{tab:effort}
\end{table*}

\section{Experiments}
\label{sec:experiments}

\subsection{Setup}
\label{sec:setup}

We evaluate twelve models from six developers: GPT-6 Sol, GPT-6 Astra, GPT-6 Luna, GPT-6.1 Sol, and GPT-6.1 Terra from OpenAI; Claude Opus 5.5, Sonnet 5.5, and Haiku 5.5 from Anthropic; Gemini 3 Flash from Google; DeepSeek Flash from DeepSeek; Kimi K2.5 from Moonshot AI; and Qwen 3.7 Flash from Alibaba.
Each model answers all 238 questions in the setting of \cref{sec:protocol}.
Model versions and settings are listed in \cref{sec:supp-models}.

\subsection{Main results}
\label{sec:main-results}

The leading agents complete most land-change analyses end to end (\cref{tab:main}; by class in \cref{sec:supp-full}).
On the 226 computable and recognition questions, Opus 5.5, GPT-6 Sol, GPT-6.1 Sol, and Sonnet 5.5 are right in all three passes on 81.0\% to 85.4\%, with overlapping intervals.
The spread across models---even within the same family---is wide: GPT-6 Luna scores 18.6 and, with Qwen 3.7 Flash, falls below the majority-class baseline of 23.9.

\begin{figure}[t]
  \centering
  \includegraphics[width=\linewidth]{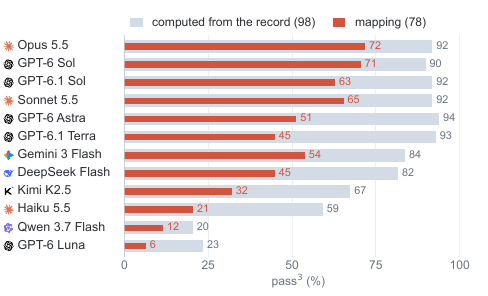}
  \caption{\textbf{Mapping from imagery is the main bottleneck.}
  pass$^3$ (\%) of each model on the 98 questions computed from variables in the record (grey) and the 78 questions that need mapping from the imagery, with references from withheld products (red).
  Models follow \cref{tab:main}, and \cref{sec:supp-reftype} gives all four reference types.}
  \label{fig:reftype}
\end{figure}

\begin{figure*}[t]
  \centering
  \includegraphics[width=\linewidth]{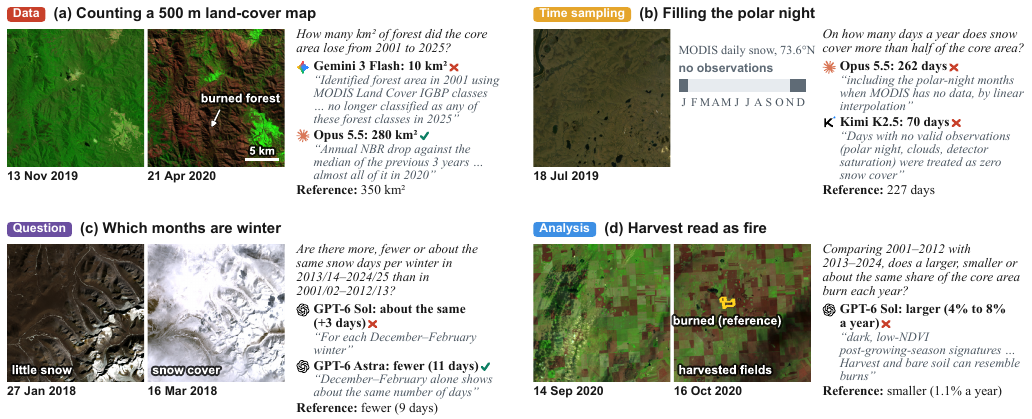}
  \caption{\textbf{Failed analyses.}
  Four runs that fail at different steps, each with an excerpt of its report, a right run where one exists, and the reference.
  Questions and quotations are shortened.
  Landsat false color (SWIR2, NIR, red) in (a) and (d) and true color elsewhere, with the same extent and stretch within each pair.
  The gap in (b) is drawn as the runs report it, and the yellow outline in (d) is the October 2020 burned area in the withheld MCD64A1 product.
  Landsat data courtesy of the U.S. Geological Survey.}
  \label{fig:failures}
\end{figure*}

\textbf{Mapping from imagery is the main bottleneck.}
The six leading models answer 92.0\% of the questions computed directly from variables in the record but only 61.1\% of those that need mapping from the imagery (\cref{fig:reftype}).
The gap holds within tasks: for the same model and task, the odds of a right answer on a mapping question are 0.31 times those on a computed one (95\% interval 0.14--0.66, \cref{sec:supp-reftype}).
Mapping is also where answers are hardest to get right and least stable: 15 of the 18 questions that no model answers right need a map, and the six leading models are right in only one or two passes on 13.7\% of mapping questions against 4.6\% of computed ones (\cref{sec:supp-consistency}).

\textbf{Some agents abstain without attempting to map.}
On the 41 recognition questions that need a map, GPT-6.1 Sol, GPT-6 Astra, and GPT-6.1 Terra answer insufficient evidence in 46\% to 67\% of their runs, against 0.9\% to 2.7\% on the 37 computable questions that need a map and at most 2\% on all computable questions (\cref{tab:effort} and \cref{sec:supp-abstention}).
Their abstaining runs make a median of two or three tool calls and mostly view no image.
The declined questions are nonetheless answerable: when these three models answer them, they are right in 85\% to 98\% of runs, and of the questions that GPT-6 Astra declines in every pass, Opus 5.5 answers nearly half right in all three.
Their lower recognition scores therefore reflect early abstention rather than wrong answers.

\textbf{When agents answer, their confidence does not always signal errors.}
Every model reports lower confidence on mapping questions than on computed ones, but by very different amounts.
GPT-6 Sol lowers its confidence by 0.11 as its accuracy falls by 13.5 points, whereas Gemini 3 Flash lowers its confidence by only 0.03 as its accuracy falls by 25.7 points, and it never declines an answerable question.
Reported confidence is therefore informative for only some agents: it separates right from wrong conclusions with an AUROC of 0.84 for Sonnet 5.5 and of 0.63 to 0.67 for Gemini 3 Flash, Qwen 3.7 Flash, and GPT-6 Luna (\cref{tab:effort} and \cref{sec:supp-confidence}).
When an agent neither declines nor lowers its confidence, as with Gemini 3 Flash, its errors are the hardest for a user to notice.

\subsection{Where analyses fail}
\label{sec:failures}

The weaker six models often fail at the first step, using the wrong data or leaving the analysis unfinished in half of their failures (\eg, \cref{fig:failures}a; \cref{sec:supp-failures}).
Such failures are rare for the leading six (7 of 72); instead, their errors arise later, in an analytic judgment: how to map a change, how to sample time, or how to define the quantity the question asks for (\cref{sec:supp-failures}).

\begin{table*}[t]
  \centering
  \footnotesize
  \setlength{\tabcolsep}{5pt}
  \renewcommand{\arraystretch}{1.05}
  \begin{tabular}{lcccccccccc}
    \toprule
    & \multicolumn{3}{c}{Computable (153)} & \multicolumn{3}{c}{Recognition (73)} & \multicolumn{3}{c}{Instrument pairs (22)} & Placebo (8) \\
    \cmidrule(lr){2-4} \cmidrule(lr){5-7} \cmidrule(lr){8-10} \cmidrule(lr){11-11}
    Model & Closed & Record & Gain & Closed & Record & Gain & Moved & Held & Beyond & Moved \\
    \midrule
    \mlogo{claude}Opus 5.5 & 62.1 & 87.6 & +25.5 & 65.8 & \textbf{80.8} & +15.1 & \textbf{7/21} & 14 & 0 & 0/7 \\
    \mlogo{openai}GPT-6 Sol & 49.0 & 86.9 & +37.9 & 57.5 & 78.1 & +20.5 & 11/19 & 7 & 1 & 0/7 \\
    \mlogo{openai}GPT-6.1 Sol & 37.9 & \textbf{89.5} & +51.6 & 60.3 & 64.4 & +4.1 & 10/20 & 9 & 1 & 0/7 \\
    \mlogo{claude}Sonnet 5.5 & 54.2 & 85.0 & +30.7 & 53.4 & 72.6 & +19.2 & 13/22 & 8 & 1 & 0/7 \\
    \mlogo{openai}GPT-6 Astra & 34.0 & 87.6 & +53.6 & 58.9 & 54.8 & $-$4.1 & 11/21 & 8 & 2 & 0/7 \\
    \mlogo{openai}GPT-6.1 Terra & 25.5 & 86.3 & +60.8 & 46.6 & 46.6 & 0.0 & 10/19 & 7 & 2 & 0/5 \\
    \mlogo{gemini}Gemini 3 Flash & 49.7 & 76.5 & +26.8 & 52.1 & 64.4 & +12.3 & 9/17 & 8 & 0 & 1/7 \\
    \mlogo{deepseek}DeepSeek Flash & 37.9 & 80.4 & +42.5 & 52.1 & 42.5 & $-$9.6 & 6/17 & 8 & \textbf{3} & 0/5 \\
    \mlogo{kimi}Kimi K2.5 & 16.3 & 62.7 & +46.4 & 30.1 & 35.6 & +5.5 & 8/15 & 4 & \textbf{3} & 2/5 \\
    \mlogo{claude}Haiku 5.5 & 25.5 & 49.7 & +24.2 & 43.8 & 23.3 & $-$20.5 & 8/13 & 4 & 1 & 3/4 \\
    \mlogo{qwen}Qwen 3.7 Flash & 4.6 & 20.9 & +16.3 & 9.6 & 16.4 & +6.8 & 7/8 & 1 & 0 & 0/1 \\
    \mlogo{openai}GPT-6 Luna & 12.4 & 17.0 & +4.6 & 27.4 & 21.9 & $-$5.5 & 7/8 & 0 & 1 & 2/4 \\
    \midrule
    \mlogoskip\textit{Majority class} & -- & 24.2 & -- & -- & 23.3 & -- & 0/6 & 6 & -- & 0/3 \\
    \mlogoskip\textit{Always no change} & -- & 18.3 & -- & -- & 4.1 & -- & 0/9 & 9 & -- & -- \\
    \bottomrule
  \end{tabular}
  \caption{\textbf{Answer sources and outcomes on instrument pairs.}
  pass$^3$ (\%) in the closed book and with the record, and the gain in points.
  Pair columns cover model--pair cases right on the original in all three passes: moved cases turn wrong after the alteration (out of those right on the original), held cases stay right and are right in the closed book or have no change as the reference, and beyond cases stay right otherwise.
  Bold marks the best model in the record (highest), moved (lowest share), and beyond (highest) columns.}
  \label{tab:analysis}
\end{table*}

A single analytic judgment can decide the answer.
For example, one question asks whether winters after 2013 have fewer snow days (\cref{fig:failures}c).
The answer depends on the definition of winter: none of the 10 runs that define it as December to February finds a decrease, whereas all 13 that use October to April or a longer period find one, as the reference does.
Among these judgments, mapping fails most often, in 31 of the 144 failures, for example when harvested fields are classified as burned (\cref{fig:failures}d).
\Cref{fig:failures}b shows a further case where observation gaps during the polar night are interpolated as snow or counted as snow-free, and \cref{sec:supp-failures} shows others such as a flood lake interpreted as a reservoir.

Moreover, the same agent often judges inconsistently across passes.
On the snow question of \cref{fig:failures}c, GPT-6.1 Sol defines winter as December to February in one pass, which yields a wrong answer, and as a longer period in the other two, which yield the right one.
Such inconsistency is common: of the 347 wrong runs of the six leading models, 121 (35\%) are on questions that the same model answers right in another pass.
These agents are thus able to answer such questions but do not apply the right judgment consistently.

\section{Prior knowledge, pixels, or the ground?}
\label{sec:analysis}

With the record, answers shift from prior knowledge to the pixels.
When the pixels are altered, however, the answers shift with them rather than reflecting change on the ground.
The question in \cref{fig:ladder}a illustrates this sequence.

\begin{figure}[t]
  \centering
  \includegraphics[width=\linewidth]{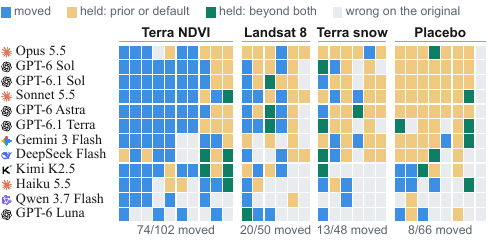}
  \caption{\textbf{Outcomes on altered instrument pairs.}
  Outcome of each model--pair case, as defined in \cref{tab:analysis}, with grey for cases wrong on the original.
  Columns are the instrument pairs grouped by the altered satellite, then the placebo pairs.}
  \label{fig:analysis}
\end{figure}

\subsection{From prior knowledge to the pixels}
\label{sec:prior-pixels}

In \cref{fig:ladder}a, the question asks whether greenness at a stable forest in northern Argentina has changed significantly.
In the closed book, GPT-6 Sol answers decrease, reasoning that clearing in the region usually lowers greenness; with the record, it computes from the pixels and finds no significant trend, in agreement with the reference.
The same shift from prior knowledge to the pixels holds for all twelve models: with the record, they score 4.6 to 60.8 points higher on computable questions (\cref{tab:analysis}), and far more answers turn from wrong to right than from right to wrong, 40.9\% of model--question pairs against 5.8\% (\cref{sec:supp-transitions}).

On recognition questions, by contrast, the record contributes less and prior knowledge persists.
With the record, four models score lower on recognition questions than in the closed book (\cref{tab:analysis}).
Even for the three leading models, recognition answers barely depend on viewing the imagery: without the image tool, they give almost the same answers (\cref{sec:supp-noimage}).
At famous areas, prior knowledge outperforms the record: the closed book scores 55.9 and the record 50.0 (\cref{sec:supp-sites}).

\subsection{To the pixels, but not yet to the ground}
\label{sec:pixels-world}

When one satellite's data change, the answers change with them.
In \cref{fig:ladder}a, Terra NDVI is scaled down from 2018 onward while nothing changes on the ground, yet GPT-6 Sol now answers decrease.
The pattern is general: 53.5\% of the cases right on the original turn wrong after the alteration (74/102 on Terra NDVI, 20/50 on Landsat 8, and 13/48 on Terra snow), 91\% of the changed answers move in the direction of the alteration, and answers on placebo pairs change in only 12\% of cases (\cref{fig:analysis} and \cref{sec:supp-pairs}).

Even among the answers that hold, most do not detect the alteration: they are already right in the closed book or have no change as the reference, and only 7.5\% of the cases right on the original hold beyond both (\cref{fig:ladder}d and \cref{fig:analysis}).
Nor does a higher score protect an answer: the eight models above 75 on computable questions still lose 33\% to 59\% of their right answers (\cref{tab:analysis}).

Agents notice the change in the pixels but interpret it as change on the ground.
On the altered pair in \cref{fig:ladder}a, one run notes that Aqua ``diverges from Terra in recent years and does not show a significant decline on its own,'' yet still answers Decrease.
Although every alteration can be found by comparison with a concurrent satellite (\cref{sec:controls}), most runs do not make even this comparison: of the wrong runs in cases that move, only 21\% report the break and 3.5\% attribute it to the instrument, while the others that report it read it as clearing, drought, or land-use change, or leave its cause open (\cref{sec:supp-coding}).

\section{Related work}
\label{sec:related}

\textbf{Remote sensing vision-language models and benchmarks.}
Remote sensing vision-language models answer questions on single scenes, in dialogues, and over image time series~\cite{kuckreja2024geochat,soni2025earthdial,irvin2025teochat}, and their benchmarks use curated imagery~\cite{an2025choice,xuan2025dynamicvl,xiao2026geommbench,fu2026omniearth}.
Robustness benchmarks test foundation models under corruptions and shifts, including a radiometric gain and a change of sensor~\cite{li2025reobench,faruk2027rspdbench,doerksen2026earthshift}.
EarthRecord evaluates decades of raw record at one site instead of a scene.

\textbf{Agents for data analysis and Earth observation.}
Sandbox agents are scored by their results in software engineering, databases, and data science~\cite{jimenez2024swebench,lei2025spider,jing2025dsbench,chen2025scienceagentbench}, and in vision, models write programs to answer visual questions~\cite{gupta2023visual,suris2023vipergpt}.
Earth observation agents call tools for retrieval, spectral indices, change detection, or disaster assessment~\cite{kao2026towards,feng2026earth,shabbir2026thinkgeo,nguyen2026terrabench,wang2026can,yu2026agenticcity}, industrial systems combine them with geospatial foundation models~\cite{bell2025earth,slonim2026google}, and Earth-Agent-Pro~\cite{lv2026earthagentpro} lets them acquire observations through tools, without controls for prior knowledge or sensor shifts.
EarthRecord gives the agent each site's multi-sensor archive as distributed, with withheld references and three passes per question.

\textbf{Controls for prior knowledge and sensor shifts.}
Models can answer visual questions without the image~\cite{jabri2016revisiting,goyal2017making,agrawal2018dont}, so benchmarks add blind controls~\cite{chen2024mmstar,yang2025thinking,fu2026omniearth} or run agents without tools or internet access~\cite{kao2026towards,feng2026earth}.
Language models hold extractable but biased geographic knowledge~\cite{manvi2024geollm,manvi2024large}, and studies measure how they weigh it against external evidence~\cite{xie2024adaptive,wu2024clasheval}.
The closed book of EarthRecord follows this line, and its instrument pairs follow paired counterfactual tests~\cite{thrush2022winoground,li2024naturalbench} and sensor changes~\cite{faruk2027rspdbench,doerksen2026earthshift} but alter the data the agent itself processes while the ground stays as it is.

\section{Conclusion}
\label{sec:conclusion}

EarthRecord evaluates whether multimodal agents can turn four decades of raw, multi-sensor satellite records into reliable land-change analysis across 56 sites and 238 questions.
Across twelve agents, the bottleneck lies not in executing code over the archive but in two scientific steps: mapping change from the imagery and verifying whether a change in the pixels reflects change on the ground.
Leading agents complete most variable-computed analyses end to end, yet falter when a question requires a map---some abstaining, mostly without viewing an image, and others judging inconsistently across passes. When one satellite's data are altered in stress-test instrument pairs, 53.5\% of right answers follow the altered sensor while only 3.5\% of failed runs blame the instrument.
Because our sandbox provides general geospatial and classical machine-learning libraries on CPU and our instrument pairs test clear-cut breaks, equipping agents with pretrained vision models for mapping and with routine cross-sensor checks for subtle drift are natural next steps.
EarthRecord and its evaluation sandbox will be released.

\section*{LLM Usage Statement}

Multimodal agents are the subjects evaluated in this benchmark (\cref{sec:experiments,sec:analysis}). In addition, large language models (Claude and GPT) were used in three supporting roles under the authors' verification:
(1)~drafting initial visual interpretations for the 37 visual-reference questions, all of which were reviewed and verified by the authors (\cref{sec:references});
(2)~coding failure steps across 144 sampled wrong runs (\cref{sec:supp-failures}) and sensor-reporting behaviors across 273 instrument-pair runs (\cref{sec:supp-coding}) from the submitted method summaries and caveats, where author checks on 36 and 31 runs respectively showed full agreement; and
(3)~assisting with analysis and plotting scripts, reference formatting, and language editing of the manuscript. All scientific claims, analyses, code, and final text were verified by the authors.

{
    \small
    \bibliographystyle{ieeenat_fullname}
    \bibliography{main}
}

\def\arxivversion{}
\clearpage
\ifdefined\arxivversion\else\setcounter{page}{1}\fi
\onecolumn
\raggedbottom
\begin{center}
  {\Large\textbf{\thetitle}\par}
  \vspace{0.5em}
  {\Large Supplementary Material\par}
  \vspace{1.0em}
\end{center}
\appendix

\section{Evaluation details}
\label{sec:supp-models}

\cref{tab:supp-models} lists the twelve models, all of which accept image input.
Each model answers the 238 questions with the record and in the closed book, and the altered versions of the 22 instrument pairs and the 8 placebo pairs with the record, three passes each.
This gives 1,518 runs per model and 18,216 in total.
Each run uses a fresh offline container with 4 CPUs and 24\,GB of memory.
Forty-six of the 238 questions are carried over from the exploratory phase in which the protocol was developed.

\begin{table}[H]
  \centering
  \footnotesize
  \setlength{\tabcolsep}{4pt}
  \begin{tabular}{lll}
    \toprule
    Model & Identifier & Effort \\
    \midrule
    GPT-6 Sol & \texttt{gpt-6-sol} & Default \\
    GPT-6 Astra & \texttt{gpt-6-astra} & Default \\
    GPT-6 Luna & \texttt{gpt-6-luna} & Default \\
    GPT-6.1 Sol & \texttt{gpt-6.1-sol} & Default \\
    GPT-6.1 Terra & \texttt{gpt-6.1-terra} & Default \\
    Claude Opus 5.5 & \texttt{claude-opus-5-5} & Default \\
    Claude Sonnet 5.5 & \texttt{claude-sonnet-5-5} & Medium \\
    Claude Haiku 5.5 & \texttt{claude-haiku-5-5} & Medium \\
    Gemini 3 Flash & \texttt{gemini-3-flash-preview} & Default \\
    DeepSeek Flash & \texttt{deepseek-flash} & Medium \\
    Kimi K2.5 & \texttt{kimi-k2.5} & Medium \\
    Qwen 3.7 Flash & \texttt{qwen3.7-flash} & Medium \\
    \bottomrule
  \end{tabular}
  \caption{\textbf{Models.} Default effort is the provider's default.}
  \label{tab:supp-models}
\end{table}

\section{Additional results of the evaluation}
\label{sec:supp-results}

The results in this section extend \cref{sec:experiments} and use the same runs and scores.

\subsection{Analysis tasks}
\label{sec:supp-tasks}

\begin{figure}[H]
  \centering
  \includegraphics[width=\linewidth]{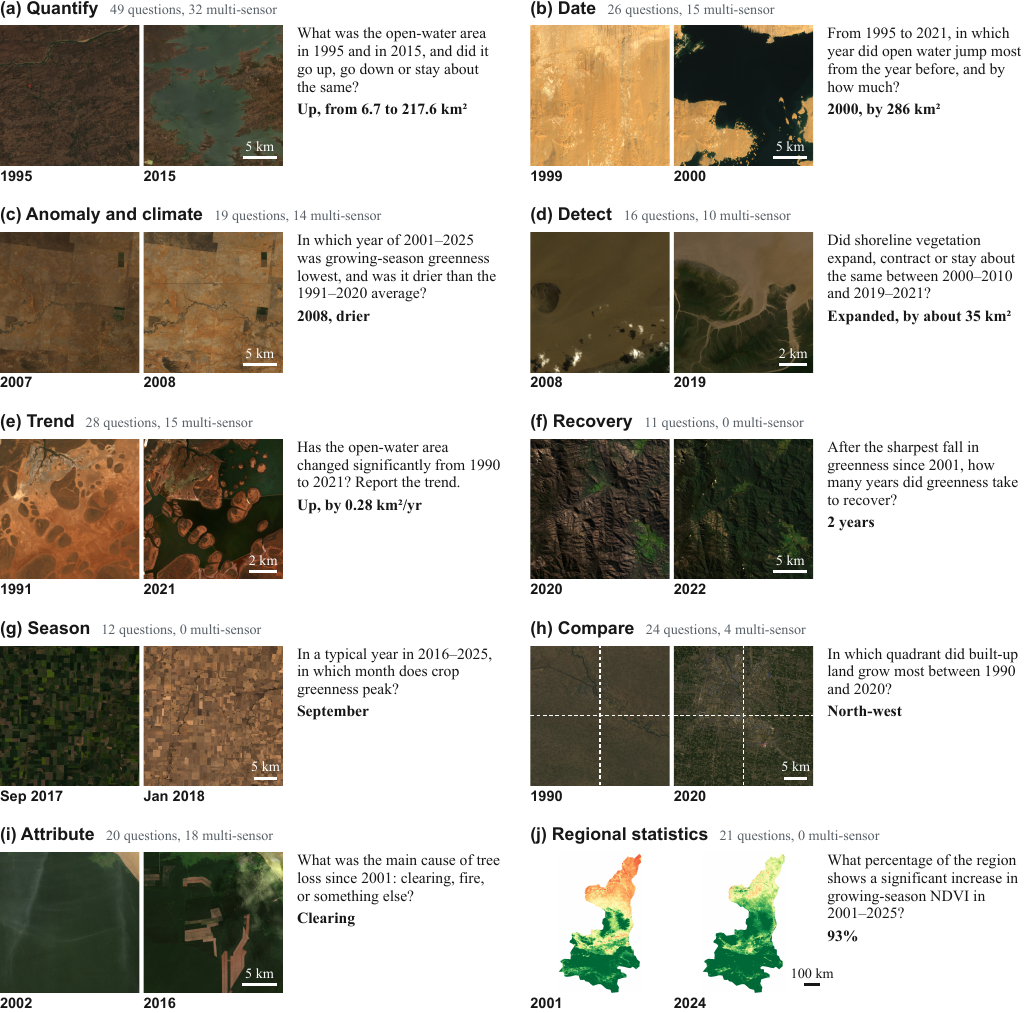}
  \caption{\textbf{Analysis tasks.}
  One shortened example question per task, with the reference answer in bold.
  Counts are over the 226 computable and recognition questions, and multi-sensor questions span two or more Landsat sensors.
  (a--i) Landsat true color, with the same extent and stretch within each pair.
  (j) MODIS NDVI of the region in late July, from 0 (red) to 0.8 (green).
  Landsat data courtesy of the U.S. Geological Survey.}
  \label{fig:supp-tasks}
\end{figure}

Each question has one analysis task, set by its question template.
\cref{fig:supp-tasks} shows one example of each, with the number of questions and of questions whose period spans two or more Landsat sensors.
In the Recovery task, 10 of the 11 questions have the reference conclusion that greenness recovered, so the majority class scores 90.9 on this task (\cref{tab:main}).

\subsection{Results by class and theme}
\label{sec:supp-themes}

\cref{tab:supp-themes} splits the overall score of \cref{tab:main} by question class and theme.
Fire is the hardest theme, with a median of 50.0 across models, and the regions are the easiest, with 96.8.

\begin{table}[H]
  \centering
  \footnotesize
  \setlength{\tabcolsep}{4.6pt}
  \begin{tabular}{lccccccccccc}
    \toprule
    & \multicolumn{2}{c}{Class} & \multicolumn{9}{c}{Theme} \\
    \cmidrule(lr){2-3} \cmidrule(lr){4-12}
    Model & Computable & Recognition & Forest & Greening & Agriculture & Water & Urban & Fire & Snow & Coast & Regions \\
    & \scriptsize(153) & \scriptsize(73) & \scriptsize(29) & \scriptsize(20) & \scriptsize(23) & \scriptsize(26) & \scriptsize(30) & \scriptsize(24) & \scriptsize(19) & \scriptsize(24) & \scriptsize(31) \\
    \midrule
    Claude Opus 5.5 & 87.6 & 80.8 & 89.7 & 75.0 & 91.3 & 76.9 & 93.3 & 66.7 & 84.2 & 83.3 & 100.0 \\
    GPT-6 Sol & 86.9 & 78.1 & 86.2 & 75.0 & 78.3 & 80.8 & 93.3 & 79.2 & 68.4 & 87.5 & 96.8 \\
    GPT-6.1 Sol & 89.5 & 64.4 & 72.4 & 75.0 & 78.3 & 88.5 & 86.7 & 62.5 & 73.7 & 87.5 & 100.0 \\
    Claude Sonnet 5.5 & 85.0 & 72.6 & 86.2 & 75.0 & 78.3 & 76.9 & 73.3 & 70.8 & 78.9 & 83.3 & 100.0 \\
    GPT-6 Astra & 87.6 & 54.8 & 69.0 & 75.0 & 78.3 & 84.6 & 76.7 & 45.8 & 78.9 & 83.3 & 96.8 \\
    GPT-6.1 Terra & 86.3 & 46.6 & 62.1 & 75.0 & 87.0 & 73.1 & 60.0 & 54.2 & 68.4 & 79.2 & 100.0 \\
    Gemini 3 Flash & 76.5 & 64.4 & 75.9 & 70.0 & 78.3 & 65.4 & 80.0 & 58.3 & 68.4 & 62.5 & 87.1 \\
    DeepSeek Flash & 80.4 & 42.5 & 69.0 & 60.0 & 73.9 & 80.8 & 46.7 & 37.5 & 68.4 & 75.0 & 96.8 \\
    Kimi K2.5 & 62.7 & 35.6 & 55.2 & 60.0 & 60.9 & 46.2 & 50.0 & 37.5 & 31.6 & 66.7 & 71.0 \\
    Claude Haiku 5.5 & 49.7 & 23.3 & 31.0 & 50.0 & 60.9 & 38.5 & 36.7 & 12.5 & 52.6 & 25.0 & 64.5 \\
    Qwen 3.7 Flash & 20.9 & 16.4 & 17.2 & 30.0 & 13.0 & 11.5 & 40.0 & 12.5 & 15.8 & 33.3 & 3.2 \\
    GPT-6 Luna & 17.0 & 21.9 & 24.1 & 30.0 & 26.1 & 11.5 & 20.0 & 12.5 & 21.1 & 25.0 & 3.2 \\
    \midrule
    \textit{Majority class} & 24.2 & 23.3 & 24.1 & 35.0 & 26.1 & 15.4 & 16.7 & 33.3 & 10.5 & 45.8 & 12.9 \\
    \textit{Always no change} & 18.3 & 4.1 & 17.2 & 20.0 & 30.4 & 7.7 & 13.3 & 4.2 & 15.8 & 4.2 & 12.9 \\
    \bottomrule
  \end{tabular}
  \caption{\textbf{Results by class and theme.} pass$^3$ (\%), with question counts in parentheses.}
  \label{tab:supp-themes}
\end{table}

\subsection{Accuracy by question class and condition}
\label{sec:supp-full}

\cref{tab:main} splits pass$^3$ with the record by analysis task.
\cref{tab:supp-full} splits the runs by question class instead, in the closed book and with the record, and gives the accuracy of single runs next to pass$^3$.
An insufficient-evidence question counts as right when the run answers insufficient evidence.
With the record, single runs of the eight leading models are right on 86.1\% to 91.5\% of computable questions and on 53.9\% to 84.0\% of recognition questions.

\begin{table}[H]
  \centering
  \footnotesize
  \setlength{\tabcolsep}{3.4pt}
  \begin{tabular}{lcccccccccc}
    \toprule
    & \multicolumn{4}{c}{Computable (153)} & \multicolumn{4}{c}{Recognition (73)} & \multicolumn{2}{c}{Insufficient (12)} \\
    \cmidrule(lr){2-5} \cmidrule(lr){6-9} \cmidrule(lr){10-11}
    & \multicolumn{2}{c}{Closed book} & \multicolumn{2}{c}{Record} & \multicolumn{2}{c}{Closed book} & \multicolumn{2}{c}{Record} & \multicolumn{2}{c}{Record} \\
    \cmidrule(lr){2-3} \cmidrule(lr){4-5} \cmidrule(lr){6-7} \cmidrule(lr){8-9} \cmidrule(lr){10-11}
    Model & One run & pass$^3$ & One run & pass$^3$ & One run & pass$^3$ & One run & pass$^3$ & One run & pass$^3$ \\
    \midrule
    Claude Opus 5.5 & 68.6 & 62.1 & 91.1 & 87.6 & 68.9 & 65.8 & 84.0 & 80.8 & 97.2 & 91.7 \\
    GPT-6 Sol & 57.1 & 49.0 & 91.1 & 86.9 & 63.9 & 57.5 & 83.6 & 78.1 & 97.2 & 91.7 \\
    GPT-6.1 Sol & 46.6 & 37.9 & 91.5 & 89.5 & 65.3 & 60.3 & 68.0 & 64.4 & 100.0 & 100.0 \\
    Claude Sonnet 5.5 & 59.7 & 54.2 & 90.2 & 85.0 & 61.6 & 53.4 & 79.9 & 72.6 & 83.3 & 75.0 \\
    GPT-6 Astra & 41.0 & 34.0 & 90.6 & 87.6 & 63.5 & 58.9 & 60.7 & 54.8 & 100.0 & 100.0 \\
    GPT-6.1 Terra & 38.1 & 25.5 & 89.8 & 86.3 & 55.7 & 46.6 & 53.9 & 46.6 & 100.0 & 100.0 \\
    Gemini 3 Flash & 59.5 & 49.7 & 86.1 & 76.5 & 60.3 & 52.1 & 69.9 & 64.4 & 13.9 & 8.3 \\
    DeepSeek Flash & 51.2 & 37.9 & 87.4 & 80.4 & 57.5 & 52.1 & 66.7 & 42.5 & 91.7 & 83.3 \\
    Kimi K2.5 & 29.6 & 16.3 & 77.1 & 62.7 & 45.2 & 30.1 & 51.1 & 35.6 & 55.6 & 50.0 \\
    Claude Haiku 5.5 & 38.6 & 25.5 & 70.2 & 49.7 & 54.8 & 43.8 & 45.2 & 23.3 & 88.9 & 83.3 \\
    Qwen 3.7 Flash & 15.3 & 4.6 & 32.0 & 20.9 & 24.2 & 9.6 & 29.2 & 16.4 & 91.7 & 75.0 \\
    GPT-6 Luna & 30.9 & 12.4 & 42.0 & 17.0 & 39.3 & 27.4 & 37.0 & 21.9 & 91.7 & 83.3 \\
    \bottomrule
  \end{tabular}
  \caption{\textbf{Accuracy by question class and condition.} Scores in \%, with question counts in parentheses.}
  \label{tab:supp-full}
\end{table}

\subsection{Results by reference type}
\label{sec:supp-reftype}

\cref{tab:supp-reftype} gives each model's score by where the reference answer comes from (\cref{sec:main-results}), and \cref{tab:supp-logit} the logistic regression behind the odds ratio in \cref{sec:main-results}.
Eleven of the twelve models score lowest on mapping questions, from 6.4 for GPT-6 Luna to 71.8 for Opus 5.5.
GPT-6 Luna scores 0.0 on the 13 record-product questions.
Relative to questions computed from variables in the record, record-product and visual questions show no significant difference.

\begin{table}[H]
  \centering
  \footnotesize
  \newcommand{\hd}[1]{\begin{tabular}[b]{@{}c@{}}#1\end{tabular}}
  \setlength{\tabcolsep}{4pt}
  \begin{tabular}{lcccc}
    \toprule
    Model & Method & \hd{Record\\product} & Mapping & Visual \\
    & \scriptsize(98) & \scriptsize(13) & \scriptsize(78) & \scriptsize(37) \\
    \midrule
    Claude Opus 5.5 & 91.8 & 100.0 & 71.8 & 91.9 \\
    GPT-6 Sol & 89.8 & 92.3 & 70.5 & 94.6 \\
    GPT-6.1 Sol & 91.8 & 92.3 & 62.8 & 89.2 \\
    Claude Sonnet 5.5 & 91.8 & 84.6 & 65.4 & 83.8 \\
    GPT-6 Astra & 93.9 & 92.3 & 51.3 & 81.1 \\
    GPT-6.1 Terra & 92.9 & 84.6 & 44.9 & 78.4 \\
    Gemini 3 Flash & 83.7 & 92.3 & 53.8 & 75.7 \\
    DeepSeek Flash & 81.6 & 100.0 & 44.9 & 70.3 \\
    Kimi K2.5 & 67.3 & 84.6 & 32.1 & 54.1 \\
    Claude Haiku 5.5 & 59.2 & 53.8 & 20.5 & 32.4 \\
    Qwen 3.7 Flash & 20.4 & 46.2 & 11.5 & 24.3 \\
    GPT-6 Luna & 23.5 & 0.0 & 6.4 & 37.8 \\
    \bottomrule
  \end{tabular}
  \caption{\textbf{Results by reference type.} pass$^3$ (\%) by the source of the reference answer, with question counts in parentheses.}
  \label{tab:supp-reftype}
\end{table}

\begin{table}[H]
  \centering
  \footnotesize
  \setlength{\tabcolsep}{4pt}
  \begin{tabular}{lccc}
    \toprule
    Term & Odds ratio & 95\% interval & $p$ \\
    \midrule
    Record product & 1.05 & 0.45--2.44 & 0.911 \\
    Mapping & 0.31 & 0.14--0.66 & 0.003 \\
    Visual & 0.65 & 0.18--2.32 & 0.510 \\
    Two or more Landsat sensors & 0.59 & 0.33--1.06 & 0.078 \\
    Period length, per decade & 0.94 & 0.71--1.24 & 0.662 \\
    \bottomrule
  \end{tabular}
  \caption{\textbf{Predictors of question difficulty.} Logistic regression of model--question pass$^3$, with model and task fixed effects and standard errors clustered by question. Reference types are relative to questions computed from record variables.}
  \label{tab:supp-logit}
\end{table}

\subsection{Consistency over passes}
\label{sec:supp-consistency}

\cref{tab:supp-consistency} counts, for each model, the 226 computable and recognition questions by the number of passes whose conclusion is right.
The six leading models are right in one or two of the three passes on 11 to 28 questions, DeepSeek Flash on 49, Haiku 5.5 on 83, and GPT-6 Luna on 100.
Mixed results are most common on mapping questions: the six leading models are right in one or two passes on 13.7\% of mapping questions and 4.6\% of questions computed from record variables, and from single runs to pass$^3$ they lose 6.8 and 2.4 points on the two groups.

\begin{table}[H]
  \centering
  \footnotesize
  \setlength{\tabcolsep}{5pt}
  \begin{tabular}{lccccc}
    \toprule
    & \multicolumn{4}{c}{Passes right} & \\
    \cmidrule(lr){2-5}
    Model & 3 & 2 & 1 & 0 & 1 or 2 \\
    \midrule
    Claude Opus 5.5 & 193 & 10 & 3 & 20 & 13 \\
    GPT-6 Sol & 190 & 11 & 9 & 16 & 20 \\
    GPT-6.1 Sol & 184 & 6 & 5 & 31 & 11 \\
    Claude Sonnet 5.5 & 183 & 12 & 16 & 15 & 28 \\
    GPT-6 Astra & 174 & 9 & 9 & 34 & 18 \\
    GPT-6.1 Terra & 166 & 12 & 8 & 40 & 20 \\
    Gemini 3 Flash & 164 & 20 & 16 & 26 & 36 \\
    DeepSeek Flash & 154 & 36 & 13 & 23 & 49 \\
    Kimi K2.5 & 122 & 38 & 24 & 42 & 62 \\
    Claude Haiku 5.5 & 93 & 59 & 24 & 50 & 83 \\
    Qwen 3.7 Flash & 44 & 26 & 27 & 129 & 53 \\
    GPT-6 Luna & 42 & 48 & 52 & 84 & 100 \\
    \bottomrule
  \end{tabular}
  \caption{\textbf{Consistency over passes.} Questions by the number of passes, out of three, with a right conclusion.}
  \label{tab:supp-consistency}
\end{table}

\subsection{Abstention}
\label{sec:supp-abstention}

Every answer field accepts insufficient evidence (\cref{sec:questions}).
\cref{tab:supp-abstention} counts it on the 12 insufficient-evidence questions, where it is the reference answer, and on the other questions, where it is scored as wrong.
Ten models abstain in at least 30 of the 36 runs on insufficient-evidence questions, Kimi K2.5 in 20, and Gemini 3 Flash in 5.
On recognition questions GPT-6.1 Sol, GPT-6 Astra, and GPT-6.1 Terra abstain in 29.7\%, 38.8\%, and 42.9\% of runs, after a median of two or three tool calls and mostly without viewing an image, whereas their answered runs on these questions take a median of five to ten calls.
Haiku 5.5, Qwen 3.7 Flash, and GPT-6 Luna abstain on computable questions as well.
On the 41 recognition questions that need a map, GPT-6.1 Sol, GPT-6 Astra, and GPT-6.1 Terra abstain in 46.3\%, 58.5\%, and 66.7\% of runs, against 8.1\% for GPT-6 Sol and 8.9\% for Sonnet 5.5.
When they answer these questions, they are right in 92.4\%, 98.0\%, and 85.4\% of runs, against 77.9\% and 75.9\%.
They decline 19, 25, and 28 recognition questions in all three passes, and Opus 5.5 answers 6, 12, and 15 of these right in all three.

\begin{table}[H]
  \centering
  \footnotesize
  \newcommand{\hd}[1]{\begin{tabular}[b]{@{}c@{}}#1\end{tabular}}
  \setlength{\tabcolsep}{2.6pt}
  \begin{tabular}{lcccccc}
    \toprule
    & \multicolumn{3}{c}{Recognition} & Comp. & \multicolumn{2}{c}{Insufficient evid.} \\
    \cmidrule(lr){2-4} \cmidrule(lr){5-5} \cmidrule(lr){6-7}
    Model & \hd{Abst.\\(\%)} & \hd{Tool\\calls} & \hd{No\\image} & \hd{Abst.\\(\%)} & \hd{Abst.\\runs} & \hd{Right\\(of 12)} \\
    \midrule
    Claude Opus 5.5 & 4.2 & 12 & 11 & 0.0 & 35 & 11 \\
    GPT-6 Sol & 4.6 & 11.5 & 80 & 0.0 & 35 & 11 \\
    GPT-6.1 Sol & 29.7 & 3 & 86 & 0.4 & 36 & 12 \\
    Claude Sonnet 5.5 & 5.0 & 13 & 82 & 0.4 & 30 & 9 \\
    GPT-6 Astra & 38.8 & 2 & 95 & 0.7 & 36 & 12 \\
    GPT-6.1 Terra & 42.9 & 3 & 98 & 2.0 & 36 & 12 \\
    Gemini 3 Flash & 0.0 & -- & -- & 0.0 & 5 & 1 \\
    DeepSeek Flash & 1.4 & 92 & 0 & 0.0 & 33 & 10 \\
    Kimi K2.5 & 4.1 & 31 & 100 & 0.0 & 20 & 6 \\
    Claude Haiku 5.5 & 32.0 & 7 & 97 & 11.8 & 32 & 10 \\
    Qwen 3.7 Flash & 39.7 & 47 & 98 & 46.8 & 33 & 9 \\
    GPT-6 Luna & 40.2 & 4 & 99 & 41.4 & 33 & 10 \\
    \bottomrule
  \end{tabular}
  \caption{\textbf{Abstention.} Runs per model that answer insufficient evidence, out of 219 recognition, 459 computable, and 36 insufficient-evidence runs. Tool calls (median) and No image (\%) refer to the abstaining recognition runs, and Right to questions right in all three passes.}
  \label{tab:supp-abstention}
\end{table}

\subsection{Reported confidence}
\label{sec:supp-confidence}

Each run reports a confidence between 0 and 1 with its answer.
\cref{tab:supp-confidence} gives, among the answered runs of each model, the AUROC of this confidence for right against wrong conclusions, its mean on each, and the accuracy at high and low confidence.
The models use the scale differently.
Gemini 3 Flash reports at least 0.9 in 89.1\% of its runs and is right in 84.9\% of them, while Opus 5.5 and Sonnet 5.5 report at least 0.9 in 1.9\% and 4.8\% of their runs and are right in all of them.
Every model reports lower confidence on mapping questions than on computed ones, by 0.02 to 0.23, while its accuracy falls by 6.0 to 27.8 points.

\begin{table}[H]
  \centering
  \footnotesize
  \setlength{\tabcolsep}{2.4pt}
  \begin{tabular}{lccccccccccc}
    \toprule
    & & \multicolumn{2}{c}{Mean} & \multicolumn{2}{c}{$\geq 0.9$} & \multicolumn{2}{c}{$< 0.7$} & \multicolumn{2}{c}{Computed (98)} & \multicolumn{2}{c}{Mapping (78)} \\
    \cmidrule(lr){3-4} \cmidrule(lr){5-6} \cmidrule(lr){7-8} \cmidrule(lr){9-10} \cmidrule(lr){11-12}
    Model & AUROC & Right & Wrong & Runs & Acc. & Runs & Acc. & Conf. & Acc. & Conf. & Acc. \\
    \midrule
    Claude Opus 5.5 & 0.79 & 0.72 & 0.57 & 1.9 & 100.0 & 37.4 & 79.0 & 0.71 & 94.2 & 0.63 & 78.8 \\
    GPT-6 Sol & 0.76 & 0.86 & 0.73 & 41.3 & 96.4 & 11.2 & 72.0 & 0.88 & 93.9 & 0.77 & 80.4 \\
    GPT-6.1 Sol & 0.79 & 0.89 & 0.80 & 54.6 & 97.6 & 4.3 & 80.8 & 0.90 & 94.5 & 0.84 & 88.6 \\
    Claude Sonnet 5.5 & 0.84 & 0.71 & 0.52 & 4.8 & 100.0 & 44.3 & 77.6 & 0.71 & 95.9 & 0.62 & 77.9 \\
    GPT-6 Astra & 0.78 & 0.87 & 0.78 & 45.4 & 98.9 & 5.6 & 81.8 & 0.88 & 94.9 & 0.82 & 87.4 \\
    GPT-6.1 Terra & 0.85 & 0.84 & 0.67 & 36.5 & 99.5 & 11.7 & 67.2 & 0.85 & 95.5 & 0.77 & 83.9 \\
    Gemini 3 Flash & 0.67 & 0.94 & 0.90 & 89.1 & 84.9 & 0.1 & 0.0 & 0.94 & 91.8 & 0.91 & 66.1 \\
    DeepSeek Flash & 0.75 & 0.78 & 0.65 & 9.7 & 100.0 & 23.3 & 77.5 & 0.76 & 93.6 & 0.72 & 78.3 \\
    Kimi K2.5 & 0.70 & 0.87 & 0.81 & 52.4 & 84.8 & 2.9 & 57.9 & 0.89 & 82.5 & 0.82 & 55.0 \\
    Claude Haiku 5.5 & 0.70 & 0.59 & 0.47 & 0.7 & 100.0 & 66.1 & 69.0 & 0.62 & 85.0 & 0.45 & 60.0 \\
    Qwen 3.7 Flash & 0.63 & 0.85 & 0.80 & 36.4 & 80.9 & 9.3 & 53.6 & 0.84 & 76.5 & 0.82 & 64.3 \\
    GPT-6 Luna & 0.67 & 0.74 & 0.60 & 20.6 & 80.5 & 35.6 & 53.5 & 0.78 & 79.2 & 0.55 & 51.4 \\
    \bottomrule
  \end{tabular}
  \caption{\textbf{Reported confidence.} Answered runs on the 226 computable and recognition questions. Runs is the share (\%) of runs in each confidence band, and Acc. the share of those that are right (\%). Computed and Mapping give the mean confidence and the accuracy (\%) on the questions computed from record variables and on those that need a map.}
  \label{tab:supp-confidence}
\end{table}

\subsection{Failed analyses}
\label{sec:supp-failures}

From the 1,232 runs on original computable and recognition questions whose conclusion is wrong, leaving out unwarranted abstentions, we drew 12 per model at random, at most one per model and question.
For each run, its method summary and caveats were read against the question and the reference to mark the first step that departs from what the question asks and explains the miss (\cref{tab:supp-failures}).
An LLM coded the failure steps; the authors checked 36 runs---14 drawn at random from the 39 marked as uncertain by the LLM and at most two per code from the rest---and agreed with all 36.

\begin{table}[H]
  \centering
  \footnotesize
  \setlength{\tabcolsep}{3pt}
  \begin{tabular}{@{}p{1.45cm}p{5.0cm}cc@{}}
    \toprule
    Step & The run & Lead & Other \\
    \midrule
    Data & counts a coarse land-cover map, uses the wrong product, misses data in the record, or joins sensors or products without making them consistent & 5 & 18 \\
    Time sampling & fills gaps such as the polar night by assumption, or rests on few or seasonally uneven scenes & 16 & 7 \\
    Question & computes a different quantity, such as winter as December to February & 15 & 6 \\
    Analysis & maps change from the imagery with a different result, judges the type or cause of a change wrongly, or reaches its conclusion through the choice of test & 27 & 15 \\
    Close call & uses a sound method whose result lies at a decision boundary & 7 & 4 \\
    Unfinished & gives no answer, or one it says it could not compute & 2 & 19 \\
    Other & reports a conclusion its own numbers contradict, or the cause is not visible & 0 & 3 \\
    \bottomrule
  \end{tabular}
  \caption{\textbf{Steps of failed analyses.}
  First wrong step of 72 failed runs of the six leading models (Lead) and 72 of the other six (Other).}
  \label{tab:supp-failures}
\end{table}

\begin{figure}[H]
  \centering
  \includegraphics[width=\linewidth]{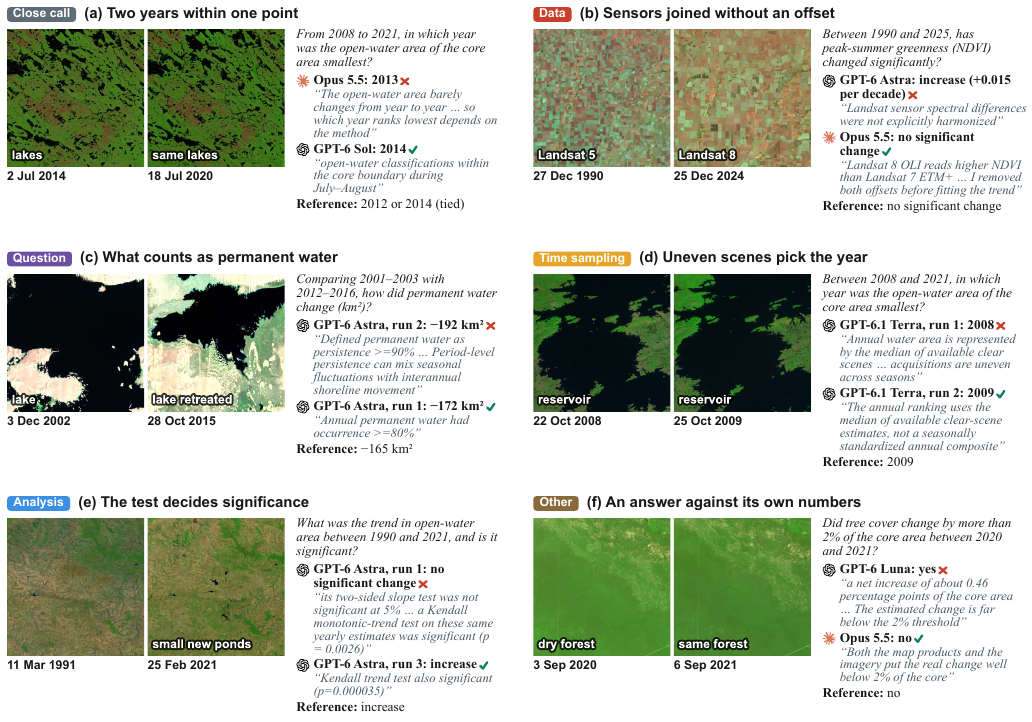}\\[2mm]
  \includegraphics[width=\linewidth]{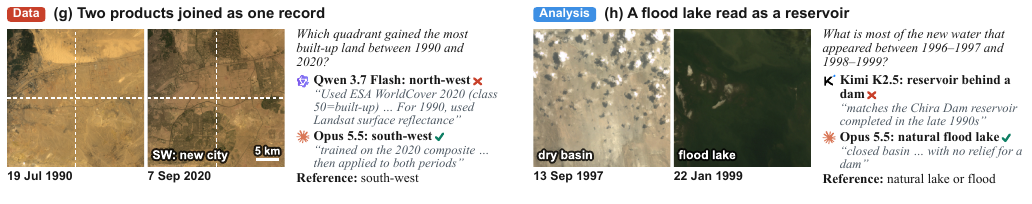}
  \caption{\textbf{More failed analyses.}
  Eight runs illustrating further failure modes from \cref{tab:supp-failures}, each with an excerpt of its report, a right run, and the reference.
  Questions and quotations are shortened.
  Landsat false color (SWIR2, NIR, red) in (a--f) and true color in (g, h), with the same extent and stretch within each pair.
  Landsat data courtesy of the U.S. Geological Survey.}
  \label{fig:supp-failures-more}
\end{figure}

\cref{fig:supp-failures-more} shows eight more failed runs, with further failure modes of \cref{tab:supp-failures} at every step except an unfinished analysis.
In three of them (\cref{fig:supp-failures-more}c--e), the right run shown for comparison is another pass of the same model, illustrating the pass-to-pass inconsistency noted in \cref{sec:failures} (121 of the 347 wrong runs of the six leading models).

\section{Additional results on answer sources and instrument pairs}
\label{sec:supp-sources}

The results in this section extend \cref{sec:analysis}.

\subsection{From the closed book to the record}
\label{sec:supp-transitions}

\cref{tab:supp-transitions} counts, for each model, the questions that turn from wrong in the closed book to right with the record and the reverse, both by pass$^3$.
On computable questions every model gains more questions than it loses.
On recognition questions GPT-6 Astra, DeepSeek Flash, Haiku 5.5, and GPT-6 Luna lose more than they gain, and GPT-6.1 Terra gains and loses 12 each.

\begin{table}[H]
  \centering
  \footnotesize
  \newcommand{\hd}[1]{\begin{tabular}[b]{@{}c@{}}#1\end{tabular}}
  \setlength{\tabcolsep}{4pt}
  \begin{tabular}{lcccccccc}
    \toprule
    & \multicolumn{4}{c}{Computable (153)} & \multicolumn{4}{c}{Recognition (73)} \\
    \cmidrule(lr){2-5} \cmidrule(lr){6-9}
    Model & \hd{Wrong to\\right} & \hd{Right to\\wrong} & \hd{Right\\in both} & \hd{Wrong\\in both} & \hd{Wrong to\\right} & \hd{Right to\\wrong} & \hd{Right\\in both} & \hd{Wrong\\in both} \\
    \midrule
    Claude Opus 5.5 & 47 & 8 & 87 & 11 & 13 & 2 & 46 & 12 \\
    GPT-6 Sol & 65 & 7 & 68 & 13 & 17 & 2 & 40 & 14 \\
    GPT-6.1 Sol & 83 & 4 & 54 & 12 & 10 & 7 & 37 & 19 \\
    Claude Sonnet 5.5 & 56 & 9 & 74 & 14 & 19 & 5 & 34 & 15 \\
    GPT-6 Astra & 84 & 2 & 50 & 17 & 9 & 12 & 31 & 21 \\
    GPT-6.1 Terra & 98 & 5 & 34 & 16 & 12 & 12 & 22 & 27 \\
    Gemini 3 Flash & 54 & 13 & 63 & 23 & 16 & 7 & 31 & 19 \\
    DeepSeek Flash & 73 & 8 & 50 & 22 & 8 & 15 & 23 & 27 \\
    Kimi K2.5 & 80 & 9 & 16 & 48 & 13 & 9 & 13 & 38 \\
    Claude Haiku 5.5 & 59 & 22 & 17 & 55 & 7 & 22 & 10 & 34 \\
    Qwen 3.7 Flash & 29 & 4 & 3 & 117 & 9 & 4 & 3 & 57 \\
    GPT-6 Luna & 23 & 16 & 3 & 111 & 4 & 8 & 12 & 49 \\
    \bottomrule
  \end{tabular}
  \caption{\textbf{From the closed book to the record.} Questions by pass$^3$ outcome in the two conditions.}
  \label{tab:supp-transitions}
\end{table}

\subsection{Without the image tool}
\label{sec:supp-noimage}

Three models answered the 73 recognition questions again without the image tool, three passes each, with everything else unchanged (\cref{tab:supp-noimage}).
Their scores change by one to four questions, and no difference is significant by a sign test.

\begin{table}[H]
  \centering
  \footnotesize
  \newcommand{\hd}[1]{\begin{tabular}[b]{@{}c@{}}#1\end{tabular}}
  \setlength{\tabcolsep}{4pt}
  \begin{tabular}{lcccccc}
    \toprule
    Model & With & Without & \hd{Closed\\book} & \hd{Only\\with} & \hd{Only\\without} & $p$ \\
    \midrule
    Claude Opus 5.5 & 59 & 55 & 48 & 6 & 2 & 0.29 \\
    GPT-6 Sol & 57 & 55 & 42 & 4 & 2 & 0.69 \\
    GPT-6.1 Sol & 47 & 46 & 44 & 2 & 1 & 1.00 \\
    \bottomrule
  \end{tabular}
  \caption{\textbf{Recognition without the image tool.}
  Questions right in all three passes, out of 73, and the sign-test $p$ value.}
  \label{tab:supp-noimage}
\end{table}

\subsection{Results by site role}
\label{sec:supp-sites}

\begin{figure}[H]
  \centering
  \includegraphics[width=\linewidth]{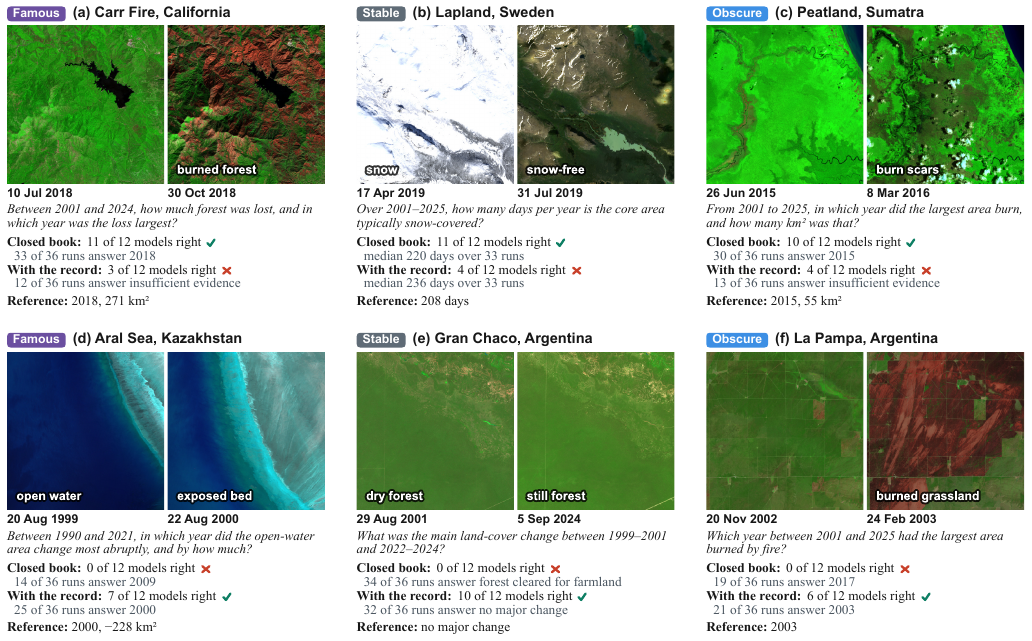}
  \caption{\textbf{Cases by site role.}
  More models are right in the closed book in the top row and with the record in the bottom row.
  Model counts are pass$^3$ over the twelve models, and run counts are out of 36, giving the most common wrong answer in the condition with fewer models right.
  Questions are shortened.
  Landsat false color (SWIR2, NIR, red) except true color in (b), with the same extent within each pair and the same stretch except in (b).
  Landsat data courtesy of the U.S. Geological Survey.}
  \label{fig:supp-sites}
\end{figure}

\cref{tab:supp-sites} pools the twelve models by the role of the site (\cref{sec:sites}).
On recognition questions at famous areas, the closed book scores 55.9 and the record 50.0.
\cref{fig:supp-sites} shows, for each role, one question where the closed book scores higher and one where the record does.
At a famous area the closed book recalls the year of the 2018 Carr Fire, while 12 of the 36 runs with the record answer insufficient evidence (\cref{fig:supp-sites}a).
For the Aral Sea it names 2009, and the record places the abrupt drop of the core area in 2000 (\cref{fig:supp-sites}d).
At a stable area in the Gran Chaco, the closed book answers forest cleared for farmland in 34 of 36 runs, and the record shows no major change (\cref{fig:supp-sites}e).
At obscure areas the closed book still recalls regional events, such as the 2015 fires of Sumatra (\cref{fig:supp-sites}c), and it places the largest fire in La Pampa in 2017 instead of 2003 (\cref{fig:supp-sites}f).

\begin{table}[H]
  \centering
  \footnotesize
  \setlength{\tabcolsep}{4pt}
  \begin{tabular}{lcccccc}
    \toprule
    & \multicolumn{3}{c}{Computable} & \multicolumn{3}{c}{Recognition} \\
    \cmidrule(lr){2-4} \cmidrule(lr){5-7}
    Site & $n$ & Closed & Record & $n$ & Closed & Record \\
    \midrule
    Famous areas & 20 & 34.2 & 77.1 & 17 & 55.9 & 50.0 \\
    Stable areas & 17 & 32.4 & 77.5 & 5 & 26.7 & 46.7 \\
    Obscure areas & 85 & 36.0 & 62.9 & 51 & 45.3 & 50.5 \\
    Regions & 31 & 29.8 & 76.6 & -- & -- & -- \\
    \bottomrule
  \end{tabular}
  \caption{\textbf{Results by site role.}
  pass$^3$ (\%) pooled over the twelve models.}
  \label{tab:supp-sites}
\end{table}

\subsection{Results by instrument pair}
\label{sec:supp-pairs}

\cref{tab:supp-pairs} lists the 30 pairs with the outcome of the model--pair cases right on the original (\cref{fig:analysis}), and how the moved answers shifted.
Of the 200 cases right on the original, 107 move, 78 hold by prior knowledge or default, and 15 hold beyond both.
On placebo pairs, 8 of 66 cases move, none of them of the six leading models.
Moved answers follow the alteration.
Of the 204 runs whose stated direction differs between the two versions, 186 move in the direction of the alteration.
A Terra NDVI scaled down turns increases into decreases or into no change, and a Terra snow cover scaled up turns decreases into no change.
The 15 cases that hold beyond both fall on six pairs, four of them on a cropland pair whose reference year, 2009, precedes the altered years, and four on a Terra snow pair.
On the 226 wrong runs of moved cases that report a confidence on both versions, the mean confidence is 0.86 on the original and 0.79 after the alteration, and it is lower in 126 of them.
Across the twelve models, the share of right cases that move and the pass$^3$ on computable questions with the record have a rank correlation of $-0.76$, driven mainly by the two weakest models, GPT-6 Luna and Qwen 3.7 Flash, which move in 7 of their 8 cases each (\cref{tab:analysis}).
Among the eight models above 75 with the record, the moved share still ranges from 33\% to 59\%, and the rank correlation weakens to $-0.36$.

\begin{table}[H]
  \centering
  \footnotesize
  \newcommand{\hd}[1]{\begin{tabular}[b]{@{}c@{}}#1\end{tabular}}
  \setlength{\tabcolsep}{2.4pt}
  \begin{tabular}{@{}llccccl@{}}
    \toprule
    Pair & Site & Right & Moved & \hd{Held:\\prior} & \hd{Held:\\beyond} & Shift in moved answers \\
    \midrule
    \multicolumn{7}{@{}l}{\textit{Terra NDVI scaled down}} \\
    \hspace{0.5em}a54-3 & cropland & 12 & 11 & 1 & 0 & no change $\rightarrow$ decrease \textcolor{gray}{19/33} \\
    \hspace{0.5em}a56-1 & greening & 11 & 11 & 0 & 0 & increase $\rightarrow$ decrease \textcolor{gray}{16/33} \\
    \hspace{0.5em}a09-1 & greening & 11 & 10 & 1 & 0 & increase $\rightarrow$ no change \textcolor{gray}{24/30} \\
    \hspace{0.5em}a33-3 & fire & 9 & 9 & 0 & 0 & 2019 $\rightarrow$ 2018 \textcolor{gray}{25/27} \\
    \hspace{0.5em}a48-3 & forest & 11 & 8 & 3 & 0 & no change $\rightarrow$ decrease \textcolor{gray}{19/24} \\
    \hspace{0.5em}a51-4 & urban & 9 & 8 & 1 & 0 & no change $\rightarrow$ decrease \textcolor{gray}{15/24} \\
    \hspace{0.5em}a45-4 & fire & 7 & 7 & 0 & 0 & recovered: yes $\rightarrow$ no \textcolor{gray}{19/21} \\
    \hspace{0.5em}a04-4 & urban & 11 & 6 & 3 & 2 & increase $\rightarrow$ decrease \textcolor{gray}{8/18} \\
    \hspace{0.5em}a31-1 & cropland & 11 & 4 & 7 & 0 & no change $\rightarrow$ decrease \textcolor{gray}{3/12} \\
    \hspace{0.5em}a39-5 & cropland & 10 & 0 & 6 & 4 &  \\
    \multicolumn{7}{@{}l}{\textit{Landsat 8 NDVI scaled down}} \\
    \hspace{0.5em}a24-1 & greening & 12 & 8 & 3 & 1 & increase $\rightarrow$ no change \textcolor{gray}{19/24} \\
    \hspace{0.5em}a41-1 & greening & 9 & 4 & 5 & 0 & no change $\rightarrow$ decrease \textcolor{gray}{2/12} \\
    \hspace{0.5em}a24-3 & greening & 9 & 1 & 5 & 3 &  \\
    \multicolumn{7}{@{}l}{\textit{Landsat 8 replaced by Landsat 7}} \\
    \hspace{0.5em}a50-1 & water & 8 & 7 & 1 & 0 & 2017 $\rightarrow$ 2002 \textcolor{gray}{6/21} \\
    \multicolumn{7}{@{}l}{\textit{Terra snow scaled up}} \\
    \hspace{0.5em}a21-1 & snow & 11 & 4 & 3 & 4 & decrease $\rightarrow$ no change \textcolor{gray}{5/12} \\
    \hspace{0.5em}a01-1 & snow & 9 & 4 & 5 & 0 & no change $\rightarrow$ increase \textcolor{gray}{6/12} \\
    \hspace{0.5em}a12-2 & snow & 10 & 2 & 8 & 0 &  \\
    \hspace{0.5em}a21-2 & snow & 4 & 2 & 1 & 1 & decrease $\rightarrow$ no change \textcolor{gray}{4/6} \\
    \hspace{0.5em}a42-1 & snow & 8 & 1 & 7 & 0 &  \\
    \hspace{0.5em}a53-3 & snow & 6 & 0 & 6 & 0 &  \\
    \multicolumn{7}{@{}l}{\textit{Landsat 8 water scaled up}} \\
    \hspace{0.5em}a06-1 & coast & 9 & 0 & 9 & 0 &  \\
    \hspace{0.5em}a20-1 & water & 3 & 0 & 3 & 0 &  \\
    \multicolumn{7}{@{}l}{\textit{Placebo: 5\% of clear scenes removed}} \\
    \hspace{0.5em}a36-1 & water & 11 & 3 & 7 & 1 &  \\
    \hspace{0.5em}a46-1 & coast & 11 & 2 & 9 & 0 &  \\
    \hspace{0.5em}a28-1 & forest & 10 & 1 & 8 & 1 &  \\
    \hspace{0.5em}a54-1 & cropland & 10 & 1 & 7 & 2 &  \\
    \hspace{0.5em}a04-1 & urban & 6 & 1 & 5 & 0 &  \\
    \hspace{0.5em}a33-1 & fire & 10 & 0 & 10 & 0 &  \\
    \hspace{0.5em}a42-2 & snow & 8 & 0 & 4 & 4 &  \\
    \hspace{0.5em}a56-2 & greening & 0 & 0 & 0 & 0 &  \\
    \bottomrule
  \end{tabular}
  \caption{\textbf{Results by instrument pair.} Right counts the models right on the original in all three passes, split as in \cref{tab:analysis}. The last column gives the reference conclusion, the most common other answer in the altered runs of the moved models, and its count over those runs, where that answer is unique and given at least twice. Site is the theme of the core area.}
  \label{tab:supp-pairs}
\end{table}

\subsection{Reports on instrument pairs}
\label{sec:supp-coding}

An LLM coded the 228 wrong runs in cases that move and the 45 runs in cases beyond both by what the method summary and caveats of each run say about the sensors: the run attributes a break in the altered period to the instrument; reports the break or a divergence from a concurrent satellite but reads it as land change or leaves its cause open; reports the altered and a concurrent satellite separately, with different results, without flagging them; gives only a generic caveat about sensors; or says nothing about them.
At the two sites with a real fire in the question period, reporting the drop caused by the fire answers the question and does not count as noticing the alteration; only a decline reported outside the burned area counts as a break.
Using two keyword passes over all 273 reports to flag candidate statements, the LLM coded each run; the authors then checked 31 runs---all 11 that the LLM marked as uncertain and 20 drawn at random within the five codes---and agreed with all 31.
Of the 228 wrong runs in cases that move, 117 say nothing about the sensors, 52 give only a generic caveat, 10 report the altered and a concurrent satellite separately without flagging the difference, 41 report the break but read it as land change or leave its cause open, and 8 attribute it to the instrument.
Of the 45 runs in cases beyond both, 10 identify the instrument and drop the affected data, six of them on the New Cairo greenness pair of \cref{tab:supp-excerpt}.

\cref{tab:supp-excerpt} sets two runs on this pair side by side.
Both report the drop in Terra NDVI in 2018.
GPT-6 Astra uses Terra alone, leaves the cause of the drop open, and answers decrease.
DeepSeek Flash compares Terra with Aqua and three other sensors, uses Terra only up to 2017, and answers increase, as the reference does.

\begin{table}[H]
  \centering
  \footnotesize
  \renewcommand{\arraystretch}{1.2}
  \begin{tabular}{@{}>{\raggedright\arraybackslash}p{2.1cm}>{\raggedright\arraybackslash}p{7.0cm}>{\raggedright\arraybackslash}p{7.0cm}@{}}
    \toprule
    & GPT-6 Astra, run 1 (moved) & DeepSeek Flash, run 1 (held beyond both) \\
    \midrule
    Data used & MODIS NDVI, Terra only & MODIS NDVI (Terra and Aqua), VIIRS, Landsat, Sentinel-2 \\
    Work & 8 tool calls, 1.5 minutes, no image viewed & 66 tool calls, 20.1 minutes, 3 images viewed \\
    Original & increase, +0.014 per decade (right) & increase, +0.014 per decade (right) \\
    Altered & decrease, $-$0.091 per decade (wrong) & increase, +0.013 per decade (right) \\
    From its report & \textit{``Used Terra MOD13Q1 throughout 2001--2025'' \ldots\ ``annual mean NDVI is approximately 0.112 in 2001, 0.131 in 2017, \mbox{-0.067} in 2018 and \mbox{-0.056} in 2025'' \ldots\ ``The cause of the break has not been independently verified.''}
    & \textit{``MODIS Terra (MOD13Q1) is unusable after 2017 in this dataset: the AOI-mean NDVI steps down by $\sim$0.20 between 2017 and 2018'' \ldots\ ``while Aqua shows no change'' \ldots\ ``sub-period fits (Aqua 2003-2019/2020, Terra 2001-2017, VIIRS 2013-2019) all give +0.011 to +0.014/decade''} \\
    \bottomrule
  \end{tabular}
  \caption{\textbf{Two runs on one altered pair.} The question asks whether greenness in New Cairo, Egypt, changed from 2001 to 2025. The altered version scales Terra NDVI down from 2018, and the reference is an increase.}
  \label{tab:supp-excerpt}
\end{table}

\end{document}